\pdfoutput=1

\documentclass[11pt]{article}

\ifx\review\undefined
\usepackage[]{acl}
\else
\usepackage[review]{acl}
\fi
\usepackage{times}
\usepackage{latexsym}

\usepackage[T1]{fontenc}
\usepackage[utf8]{inputenc}
\usepackage{xspace}

\usepackage{mathtools}
\usepackage{amssymb}
\usepackage{amsthm}
\usepackage{bm}  %
\usepackage{dsfont}
\usepackage{cancel}
\usepackage{xfrac}
\usepackage{siunitx}

\usepackage[margin=10pt,font=small,labelfont=bf,labelsep=endash]{caption}    %
\usepackage{subcaption}
\usepackage{mdframed}
\usepackage{tikz}
\usetikzlibrary{calc}

\usepackage{calc}

\usepackage[english]{babel}
\usepackage{microtype}
\usepackage{todonotes}
\usepackage{blindtext}
\usepackage{booktabs}
\usepackage{etoc}
\usepackage{enumitem}
\setlist[itemize]{topsep=5pt,parsep=0pt,partopsep=0pt,itemsep=5pt} %
\setlist[enumerate]{topsep=5pt,parsep=0pt,partopsep=0pt,itemsep=5pt} %

\usepackage[noend]{algpseudocode}    %
\usepackage[algochapter]{algorithm}  %

\DeclareMathOperator{\Normal}{\mathcal{N}}
\DeclareMathOperator{\AV}{AV}
\DeclareMathOperator{\AU}{AU}
\DeclareMathOperator{\TA}{TA}
\DeclareMathOperator{\BR}{BR}
\DeclareMathOperator{\LR}{LR}
\DeclareMathOperator{\dom}{\!dom}

\newcommand{\E}{\mathbb{E}}

\newcommand{\A}{\mathcal{A}}

\newcommand{\lagom}{L\textsuperscript{\hspace{-3pt}A}G\textsubscript{O}M$\boldsymbol{\cdot}$NLP}
\newcommand{\als}{ALs\xspace}
\newcommand{\fls}{FLs\xspace}

\newcommand{\hypo}[1]{\textbf{#1}}
\newcommand{\Hone}{\hypo{H1}\xspace}
\newcommand{\Htwo}{\hypo{H2}\xspace}
\newcommand{\Hthree}{\hypo{H3}\xspace}

\newcommand\scalemath[2]{\scalebox{#1}{\mbox{\ensuremath{\displaystyle #2}}}}

\let\originalleft\left
\let\originalright\right
\renewcommand{\left}{\mathopen{}\mathclose\bgroup\originalleft}
\renewcommand{\right}{\aftergroup\egroup\originalright}

\makeatletter  %
\def\fixedsectionspace{\kern1pt\@gobble}
\def\fixedeqspace{\kern1.25pt\@gobble}
\makeatother

\newenvironment{linequote}{%
    \begin{mdframed}[
		linewidth=1.5pt,
		topline=false,
		rightline=false,
		bottomline=false,
		leftmargin = 0.1em,
		rightmargin=-0.75em,
		usetwoside=false,
	]
	\small\slshape 
}{%
    \end{mdframed}
}

\newcommand\nonumberfootnote[1]{%
  \begingroup
  \renewcommand\thefootnote{}\footnote{#1}%
  \addtocounter{footnote}{-1}%
  \endgroup
}

\definecolor{highlightblue}{HTML}{4C76B3}
\definecolor{backgroundblue}{HTML}{DAE8FC}
\definecolor{highlightgreen}{HTML}{589D54}
\definecolor{backgroundgreen}{HTML}{D5E8D4}
\definecolor{highlightred}{HTML}{9F1010}

\definecolor{highlightorange}{HTML}{F68E23}

\definecolor{high}{HTML}{50BD4A}
\definecolor{mid}{HTML}{F7E379}
\definecolor{low}{HTML}{EF5F4A}

\definecolor{ghigh}{HTML}{7491C2}
\definecolor{gmid}{HTML}{A2B5D6}
\definecolor{glow}{HTML}{C6D2E6}

\definecolor{bhigh}{HTML}{FF7C7C}
\definecolor{bmid}{HTML}{FFA4A4}
\definecolor{blow}{HTML}{FFC9C9}

\definecolor{whigh}{HTML}{76B472}
\definecolor{wmid}{HTML}{A4CDA2}
\definecolor{wlow}{HTML}{E3F0E2}

\definecolor{topcolor}{HTML}{D9D979}
\definecolor{bottomcolor}{HTML}{FFCF9F}

\newcommand{\sep}{\textcolor{black}{\textbf{/}}}
\newcommand{\stem}[1]{\textcolor{highlightgreen}{#1}}
\newcommand{\sufone}[1]{\textcolor{highlightred}{#1}}
\newcommand{\suftwo}[1]{\textcolor{highlightorange}{#1}}
\newcommand{\sufthree}[1]{\textcolor{highlightblue}{#1}}

\newcommand{\toksep}{\textcolor{black}{~\textvisiblespace~}}

\usepackage{makecell}

\usepackage{graphicx}

\usepackage{multirow}
\usepackage{siunitx}

\usepackage{hyperref}

\usepackage[table]{xcolor}
\usepackage{arydshln}
\usepackage{hhline}  %

\usepackage{etoolbox}
\usepackage{pgf}
\usepackage{xargs}

\newcommand*{\opacity}{80}

\newcommandx{\tgrad}[4][1=0.0, 2=0.5, 3=1.0]{%
    \ifdim #4 pt > #2 pt%
        \pgfmathparse{max(min(100.0*(#4-#2)/(#3-#2),100.0),0)}%
        \xdef\PercentColor{\pgfmathresult}%
        \cellcolor{high!\PercentColor!mid!\opacity}#4%
    \else
        \pgfmathparse{max(min(100.0*(#2-#4)/(#2-#1),100.0),0)}%
        \xdef\PercentColor{\pgfmathresult}%
        \cellcolor{low!\PercentColor!mid!\opacity}#4%
    \fi
}

\newcommandx{\gradproxy}[4][1=0.0, 2=0.5, 3=1.0]{%
    \ifdim #4 pt > #2 pt%
        \pgfmathparse{max(min(100.0*(#4-#2)/(#3-#2),100.0),0)}%
        \xdef\PercentColor{\pgfmathresult}%
        \cellcolor{ghigh!\PercentColor!gmid!\opacity}#4%
    \else
        \pgfmathparse{max(min(100.0*(#2-#4)/(#2-#1),100.0),0)}%
        \xdef\PercentColor{\pgfmathresult}%
        \cellcolor{glow!\PercentColor!gmid!\opacity}#4%
    \fi
}

\newcommandx{\badproxy}[4][1=0.0, 2=0.5, 3=1.0]{%
    \ifdim #4 pt > #2 pt%
        \pgfmathparse{max(min(100.0*(#4-#2)/(#3-#2),100.0),0)}%
        \xdef\PercentColor{\pgfmathresult}%
        \cellcolor{bhigh!\PercentColor!bmid!\opacity}#4%
    \else
        \pgfmathparse{max(min(100.0*(#2-#4)/(#2-#1),100.0),0)}%
        \xdef\PercentColor{\pgfmathresult}%
        \cellcolor{blow!\PercentColor!bmid!\opacity}#4%
    \fi
}

\newcommandx{\wordproxy}[4][1=0.0, 2=0.5, 3=1.0]{%
    \ifdim #4 pt > #2 pt%
        \pgfmathparse{max(min(100.0*(#4-#2)/(#3-#2),100.0),0)}%
        \xdef\PercentColor{\pgfmathresult}%
        \cellcolor{whigh!\PercentColor!wmid!\opacity}#4%
    \else
        \pgfmathparse{max(min(100.0*(#2-#4)/(#2-#1),100.0),0)}%
        \xdef\PercentColor{\pgfmathresult}%
        \cellcolor{wlow!\PercentColor!wmid!\opacity}#4%
    \fi
}

\newcommandx{\tophl}[1]{%
    \cellcolor{topcolor!\opacity}#1%
}
\newcommandx{\bottomhl}[1]{%
    \cellcolor{bottomcolor!\opacity}#1%
}

\usepackage{collcell}
\newcolumntype{R}{>{\collectcell\tgrad}c<{\endcollectcell}}

\title{Confounding Factors in Relating Model Performance to Morphology}
\author{
    Wessel Poelman${}^*$ \and Thomas Bauwens${}^*$ \and Miryam de Lhoneux \\
    \vspace{-0.9em} \\
    \lagom, Department\ of Computer Science, KU Leuven \\
    \texttt{firstname.lastname@kuleuven.be}
}

\begin{document}
\maketitle
\frenchspacing
\ifx\anon\undefined
\nonumberfootnote{* Equal contribution.}%
\fi
\begin{abstract}
	The extent to which individual language characteristics influence tokenization and language modeling is an open question.
Differences in morphological systems have been suggested as both unimportant and crucial to consider \cite[\emph{inter alia}]{cotterell2018are,gerz2018language,park2021morphology}.
We argue this conflicting evidence is due to confounding factors in experimental setups, making it hard to compare results and draw conclusions.
We identify such factors in analyses trying to answer the question of \emph{whether, and how, morphology relates to language modeling.}
Next, we re-assess three hypotheses by \citet{arnett2025why} for why modeling agglutinative languages results in higher perplexities than fusional languages: they look at morphological alignment of tokenization, tokenization efficiency, and dataset size.
We show that each conclusion includes confounding factors and suggest methodological improvements.
Finally, we introduce token bigram metrics as an intrinsic way to predict the difficulty of causal language modeling, and find that they are \emph{gradient proxies} for morphological complexity that do not require expert annotation. %
Ultimately, we outline necessities to reliably answer whether, and how, morphology relates to language modeling.

\end{abstract}
\vspace*{0pt}

\section{Introduction}\label{sec:intro}
Are certain languages \emph{inherently} easier or harder to model \cite{cotterell2018are,mielke2019what}? The interplay between language modeling and individual differences among languages is an open problem. One angle of approach is morphological complexity \cite{gerz2018language,park2021morphology}: if the internal structure of words is more unpredictable in one language than another according to some standard, then perhaps language models (LMs) have a harder time learning to predict text in that language.
Morphological systems are widely recognized as being gradient, but coarse groupings are often used, especially in NLP \cite{oncevay2022quantifying,amrhein2021how}.
\emph{Agglutinative} languages (\als) tend to add one grammatical feature to a word with each added morpheme, 
resulting in long words with many morphemes. %
\emph{Fusional} languages (\fls) tend to express information through inflection, where a single morpheme can express multiple features, resulting in shorter words with fewer morphemes.
There is particular focus in NLP to contrast \als and \fls.
Results have been mixed, with some evidence pointing to \als being harder to model than \fls \cite[\textit{e.g.,}][]{cotterell2018are,gerz2018language,gerz2018relationa} whereas others have shown that there is no difference between the two \cite[\textit{e.g.,}][]{mielke2019what,arnett2025why}.

\begin{figure}[t]
    \centering
    \begin{mdframed}[linewidth=0pt,backgroundcolor=black!2.5]
    \includegraphics[width=\columnwidth]{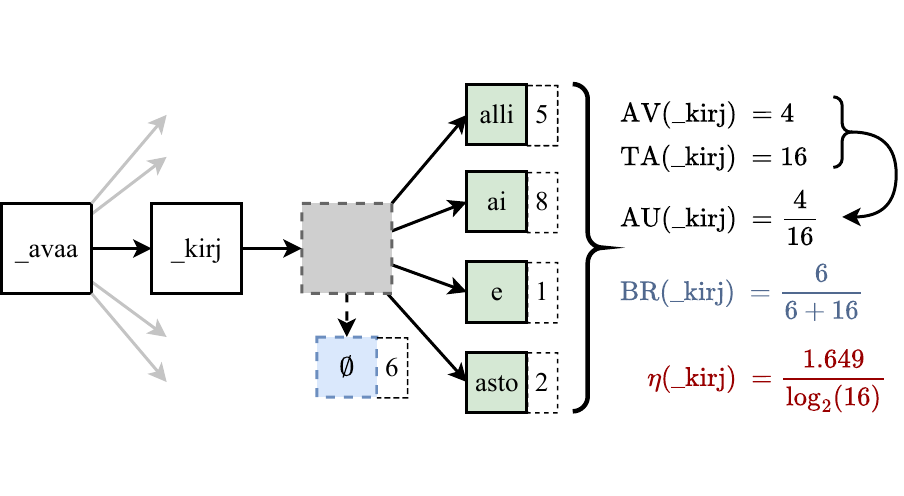}
    \end{mdframed}
    \caption{Computation of our tokenizer-based gradient proxies of morphology (\autoref{sec:metrics}) for the right accessors of a Finnish subword \texttt{\_kirj}: \textsl{accessor variety} (AV), 
    \textsl{total accessors} (TA),
    \textsl{uniqueness} (AU), and 
    \textsl{entropic efficiency} ($\eta$).
    The metrics are computed in fixed-size windows for each subword in the vocabulary, to mimic MATTR by \citet{covington_cutting_2010}. Our metrics better capture the relation between morphology and tokenization compared to word-based or unigram evaluation metrics.\vspace{-1em}}
    \label{fig:av}
\end{figure}
Unfortunately, such analyses often introduce confounding factors when studying \emph{whether, and how, morphology relates to language modeling.}
A recent, award-winning entry in the list of evidence is a study by \citet{arnett2025why}, who propose three hypotheses for why there might be a gap in perplexity (PPL) of monolingual causal LMs (CLMs) trained on \als versus \fls:

\begin{itemize}
    \item \textbf{H1:} Subword tokenization is less morphologically aligned for \als.
    \item \textbf{H2:} Subword vocabularies are used more inefficiently for \als ("worse tokenizer quality").
    \item \textbf{H3:} Less training data is available for \als.
\end{itemize}
It is sometimes wrongly assumed that subword tokenization segments words into morphemes \citep{bostrom2020byte} and the effects of this on tokenization and language modeling are unclear, hence \textbf{H1}.
Intrinsic tokenizer evaluations, such as measuring vocabulary distributions, could reveal differences between languages that might lead us to an explanation for the performance gap.
Combining these corpus-based metrics with insights from morphology might prove useful, thus \textbf{H2}. 
And lastly, \textbf{H3} is less directly related to morphology, but provides a good alternative to "just" word formation strategies as an explanation.
Hypotheses like these are proposed regularly in such analyses and \citeauthor{arnett2025why} summarize them well.

Ultimately, we outline what experimental conditions and metrics are necessary to reliably answer the aforementioned question.
Our contributions:
\begin{itemize}
    \item We list confounding factors that have to be taken into account when attempting to answer the central question above. They can be seen as criteria for an "ideal" experiment.  %
    \item We take the aforementioned hypotheses, identify confounding factors, and suggest methodological improvements to control for them.
    \item We propose predicting CLM difficulty with the variety and entropic efficiency of neighboring tokens, and find that they are proxies for morphological complexity.
\end{itemize}

\section{Related Work}\label{sec:related}
\paragraph{Language Modeling and Morphology.}
The relation between morphology and language models is mainly approached in two ways. The first asks \emph{whether certain languages are harder to model} \cite[\textit{inter alia}]{cotterell2018are,mielke2019what,park2021morphology,gerz2018language,gerz2018relationa}.
Studies taking this approach emphasize the use of parallel data, focus on model evaluations, and try to keep most experimental variables constant, except for the languages investigated.
The second way asks \emph{how aligned model architectures and tokenizers are with morphology} \cite[\textit{inter alia}]{bostrom2020byte,amrhein2021how,bauwens2024bpeknockout,limisiewicz2024myteb}.
Larger, unaligned corpora are often used, tokenizers are the main evaluation target (using reference lexicons or segmentations), and attention is paid to specific word-formation mechanisms (e.g.\ compounding).

These approaches can be at odds with each other; parallel corpora are generally quite small and specific treatments introduce potential confounds.
The previously mentioned study by \citet{arnett2025why} attempts to combine these perspectives.
We discuss their findings in \autoref{sec:hypotheses}.

\paragraph{Morphological Complexity.}
The "complexity" of a language can refer to many aspects of it \cite{sampson2009language}. Since a CLM builds words by predicting tokens, we care about the complexity of word formation, i.e.\ morphology.

Complexity metrics can be corpus statistics that measure how a language uses a vocabulary of characters/subwords/words. 
These include \emph{mean word length (MWL)}, \emph{type-token ratio (TTR)}, and \emph{moving-average TTR (MATTR)} \cite{kettunen2014can,bentz2016comparison,park2021morphology,coltekin2023what}. These metrics somewhat relate to morphology since \als tend to have more unique words that use more morphemes, resulting in higher values for the aforementioned metrics.
They have been regularly used as gradient proxies of morphology.

Other metrics are corpus-agnostic and make use of expert annotations or grammars; in their simplest form, they may be binary typological groupings \cite{wal20}. 
One can also quantify the richness of paradigms. 
Either by counting the number of paradigms or by counting how many word forms paradigms can produce on average (both referred to as \emph{E-complexity}), or by calculating how predictable other forms of a paradigm become once one is known
\citep[conditional entropy, known as \emph{I-complexity};][]{ackerman2013morphological,cotterell2018are}. 
These metrics do not reflect how a vocabulary encodes a text, which is what happens in language modeling.

In \autoref{sec:metrics}, we introduce metrics combining both perspectives: they are \emph{corpus-based} to have a direct connection with tokenization and avoid the need for experts, but, unlike for instance MWL or TTR, look at relations \emph{between}\footnote{Just as I-complexity provides more nuance than E-complexity by relating suffices \emph{to each other}.} tokens rather than tokens in isolation.
Additionally, by using \emph{tokens} instead of \emph{words}, we calculate the proxy using the same units a language model would generally use.

\section{Confounding Factors}\label{sec:factors}
It is not obvious how morphology impacts language modeling. What is clear is that research that seeks to draw reliable conclusions relating the two must control for the following confounding factors:
\begin{enumerate}
    \item \textbf{Languages}: What set of languages is under consideration? If multiple hypotheses are tested, that set should ideally stay constant.
    \item \textbf{Grouping}: If results/languages are grouped, is there enough in-group agreement to justify this?
    Coarse morphological groupings hide potentially relevant information (\autoref{tab:europarl}).
    \item \textbf{Tokenization algorithm}: What subword tokenization algorithm is used? What are its hyperparameters? The vocabulary of units and segmentation choice both influence the entire pipeline and conclusions drawn. ULM \citep{kudo2018subword} is more morphologically aligned than BPE \citep{sennrich2016neural} for e.g.\ English and Japanese \citep{bostrom2020byte}.
    \item \textbf{Vocabulary size vs.\ data size}: How does the amount of subword types relate to the amount of training data? If a tokenizer's training corpus is too small relative to the vocabulary size, it will partially store a long tail of corpus-specific strings \citep{reddy_how_2025}. If the language model's corpus is too small relative to the vocabulary size, it will have poorly trained embeddings \citep{rumbelow_solidgoldmagikarp_2023}.\footnote{Empirically, \citet{ding_call_2019} found that transformer models prefer smaller vocabularies than often used.} 
    \item \textbf{Corpus domain}: Are tokenizers and models trained on the same data? Are datasets comparable across languages (ideally multi-parallel or similar amounts of data)?
    \item \textbf{Performance indicator}: What metric is used to evaluate and compare tokenizers and models across languages? Is the setup monolingual or multilingual? Is the metric comparable between any two languages? In \autoref{sec:ppl}, we argue against comparing monolingual PPLs of different tokenizers and test texts.
\end{enumerate}
These factors show a way \emph{towards} an ideal experimental setup.\footnote{There are more factors not directly relevant, see \autoref{sec:limitations}.}
Practically, one must work \emph{backwards} from this ideal to a feasible setup in terms of data and analysis.
We now re-assess the hypotheses of \citeauthor{arnett2025why} and outline broader concerns.

\section{Re-assessment of Hypotheses}\label{sec:hypotheses}

\subsection{H1: Morphological Misalignment}\label{sec:h1}
A segmentation of a word is said to be morphologically aligned when the splits it places match the boundaries between the morphs (i.e.\ the visible parts of morphemes) that make up the word \citep{kurimo_unsupervised_2006}. Misalignment means oversegmentation of a morph and/or a token containing characters from more than one morph.

The argument for better morphological alignment causing better language modeling is that models cannot see characters.
When a morph's characters are scattered across tokens, a model may struggle to know from their embeddings that the morph is there, potentially missing its semantics.

\begin{table*}[t]
	\centering
	\resizebox{0.9\linewidth}{!}{\begin{tabular}{l||cccccccccccc}
	           & \multicolumn{3}{c|}{Full}                                                                              & \multicolumn{3}{c|}{Full${}_{\geq\!3}$ (stem-suffix)}                                                                              & \multicolumn{3}{c|}{Full${}_{\geq\!3}$ (suffix-suffix)}                                                                              & \multicolumn{3}{c}{MorphScore}                                                        \\
	           & Pr                          & Re                        & \multicolumn{1}{c|}{$F_1$}                     & Pr                                                & Re                        & \multicolumn{1}{c|}{$F_1$}                     & Pr                                               & Re                        & \multicolumn{1}{c|}{$F_1$}                     & Pr                             & \underline{\textbf{Re}}                        & $F_1$                    \\\hhline{*{13}{=}}
	German     & \tgrad[0][50][100]{28.27}   & \tgrad[0][50][100]{61.64} & \multicolumn{1}{c|}{\tgrad[0][50][100]{38.76}} & \tgrad[0][50][100]{31.85}                         & \tgrad[0][50][100]{88.68} & \multicolumn{1}{c|}{\tgrad[0][50][100]{46.87}} & \tgrad[0][50][100]{6.43}                         & \tgrad[0][50][100]{17.89} & \multicolumn{1}{c|}{\tgrad[0][50][100]{9.46}}  & \cellcolor{black!10}           & \cellcolor{black!10}      & \cellcolor{black!10}     \\
    English    & \tgrad[0][50][100]{29.65}   & \tgrad[0][50][100]{62.44} & \multicolumn{1}{c|}{\tgrad[0][50][100]{40.21}} & \tgrad[0][50][100]{32.88}                         & \tgrad[0][50][100]{86.18} & \multicolumn{1}{c|}{\tgrad[0][50][100]{47.60}} & \tgrad[0][50][100]{17.85}                        & \tgrad[0][50][100]{44.82} & \multicolumn{1}{c|}{\tgrad[0][50][100]{25.53}} & \tgrad[0][50][100]{65.88}      & \tgrad[0][50][100]{20.85} & \tgrad[0][50][100]{31.67}\\
	Polish     & \tgrad[0][50][100]{23.91}   & \tgrad[0][50][100]{41.29} & \multicolumn{1}{c|}{\tgrad[0][50][100]{30.28}} & \tgrad[0][50][100]{32.79}                         & \tgrad[0][50][100]{57.14} & \multicolumn{1}{c|}{\tgrad[0][50][100]{41.67}} & \tgrad[0][50][100]{31.15}                        & \tgrad[0][50][100]{52.78} & \multicolumn{1}{c|}{\tgrad[0][50][100]{39.18}} & \cellcolor{black!10}           & \cellcolor{black!10}      & \cellcolor{black!10}     \\
	Swedish    & \tgrad[0][50][100]{43.81}   & \tgrad[0][50][100]{42.99} & \multicolumn{1}{c|}{\tgrad[0][50][100]{43.40}} & \tgrad[0][50][100]{27.58}                         & \tgrad[0][50][100]{48.58} & \multicolumn{1}{c|}{\tgrad[0][50][100]{35.19}} & \tgrad[0][50][100]{23.44}                        & \tgrad[0][50][100]{34.60} & \multicolumn{1}{c|}{\tgrad[0][50][100]{27.95}} & \cellcolor{black!10}           & \cellcolor{black!10}      & \cellcolor{black!10}     \\
    Russian    & \tgrad[0][50][100]{25.32}   & \tgrad[0][50][100]{30.78} & \multicolumn{1}{c|}{\tgrad[0][50][100]{27.78}} & \tgrad[0][50][100]{19.71}                         & \tgrad[0][50][100]{44.33} & \multicolumn{1}{c|}{\tgrad[0][50][100]{27.29}} & \tgrad[0][50][100]{9.22}                         & \tgrad[0][50][100]{15.13} & \multicolumn{1}{c|}{\tgrad[0][50][100]{11.46}} & \cellcolor{black!10}           & \cellcolor{black!10}      & \cellcolor{black!10}     \\
    Catalan    & \tgrad[0][50][100]{29.58}   & \tgrad[0][50][100]{32.37} & \multicolumn{1}{c|}{\tgrad[0][50][100]{30.92}} & \tgrad[0][50][100]{23.40}                         & \tgrad[0][50][100]{39.58} & \multicolumn{1}{c|}{\tgrad[0][50][100]{29.42}} & \tgrad[0][50][100]{11.60}                        & \tgrad[0][50][100]{19.62} & \multicolumn{1}{c|}{\tgrad[0][50][100]{14.58}} & \cellcolor{black!10}           & \cellcolor{black!10}      & \cellcolor{black!10}     \\
    Spanish    & \tgrad[0][50][100]{27.08}   & \tgrad[0][50][100]{41.46} & \multicolumn{1}{c|}{\tgrad[0][50][100]{32.76}} & \tgrad[0][50][100]{13.65}                         & \tgrad[0][50][100]{38.18} & \multicolumn{1}{c|}{\tgrad[0][50][100]{20.11}} & \tgrad[0][50][100]{17.33}                        & \tgrad[0][50][100]{47.82} & \multicolumn{1}{c|}{\tgrad[0][50][100]{25.44}} & \tgrad[0][50][100]{52.50}      & \tgrad[0][50][100]{34.15} & \tgrad[0][50][100]{41.38}\\
    Czech      & \tgrad[0][50][100]{20.98}   & \tgrad[0][50][100]{35.31} & \multicolumn{1}{c|}{\tgrad[0][50][100]{26.32}} & \tgrad[0][50][100]{14.24}                         & \tgrad[0][50][100]{27.44} & \multicolumn{1}{c|}{\tgrad[0][50][100]{18.75}} & \tgrad[0][50][100]{17.69}                        & \tgrad[0][50][100]{34.06} & \multicolumn{1}{c|}{\tgrad[0][50][100]{23.29}} & \cellcolor{black!10}           & \cellcolor{black!10}      & \cellcolor{black!10}     \\
    French     & \tgrad[0][50][100]{23.51}   & \tgrad[0][50][100]{41.89} & \multicolumn{1}{c|}{\tgrad[0][50][100]{30.12}} & \tgrad[0][50][100]{14.89}                         & \tgrad[0][50][100]{25.88} & \multicolumn{1}{c|}{\tgrad[0][50][100]{18.91}} & \tgrad[0][50][100]{19.43}                        & \tgrad[0][50][100]{27.39} & \multicolumn{1}{c|}{\tgrad[0][50][100]{22.73}} & \cellcolor{black!10}           & \cellcolor{black!10}      & \cellcolor{black!10}     \\
    Portuguese & \tgrad[0][50][100]{12.43}   & \tgrad[0][50][100]{22.39} & \multicolumn{1}{c|}{\tgrad[0][50][100]{15.98}} & \tgrad[0][50][100]{9.52}                          & \tgrad[0][50][100]{14.75} & \multicolumn{1}{c|}{\tgrad[0][50][100]{11.57}} & \tgrad[0][50][100]{10.78}                        & \tgrad[0][50][100]{12.54} & \multicolumn{1}{c|}{\tgrad[0][50][100]{11.60}} & \cellcolor{black!10}           & \cellcolor{black!10}      & \cellcolor{black!10}     \\ \hline
	Hungarian  & \tgrad[0][50][100]{47.22}   & \tgrad[0][50][100]{69.58} & \multicolumn{1}{c|}{\tgrad[0][50][100]{56.26}} & \tgrad[0][50][100]{35.17}                         & \tgrad[0][50][100]{71.20} & \multicolumn{1}{c|}{\tgrad[0][50][100]{47.08}} & \tgrad[0][50][100]{21.85}                        & \tgrad[0][50][100]{44.21} & \multicolumn{1}{c|}{\tgrad[0][50][100]{29.24}} & \tgrad[0][50][100]{57.59}      & \tgrad[0][50][100]{43.05} & \tgrad[0][50][100]{49.27}\\
	Finnish    & \tgrad[0][50][100]{19.43}   & \tgrad[0][50][100]{35.35} & \multicolumn{1}{c|}{\tgrad[0][50][100]{25.08}} & \tgrad[0][50][100]{14.10}                         & \tgrad[0][50][100]{37.20} & \multicolumn{1}{c|}{\tgrad[0][50][100]{20.45}} & \tgrad[0][50][100]{7.52}                         & \tgrad[0][50][100]{19.85} & \multicolumn{1}{c|}{\tgrad[0][50][100]{10.91}} & \cellcolor{black!10}           & \cellcolor{black!10}      & \cellcolor{black!10}  \\
	Turkish    & \tgrad[0][50][100]{74.34}   & \tgrad[0][50][100]{34.25} & \multicolumn{1}{c|}{\tgrad[0][50][100]{46.90}} & \tgrad[0][50][100]{32.50}                         & \tgrad[0][50][100]{33.10} & \multicolumn{1}{c|}{\tgrad[0][50][100]{32.80}} & \tgrad[0][50][100]{49.58}                        & \tgrad[0][50][100]{28.31} & \multicolumn{1}{c|}{\tgrad[0][50][100]{36.04}} & \tgrad[0][50][100]{23.18}      & \tgrad[0][50][100]{48.27} & \tgrad[0][50][100]{31.32} 
\end{tabular}
}
	\caption{Morphological boundary recognition. Full segmentations are from MorphyNet \cite{batsuren2021morphynet} and MorphoChallenge \cite[][only Turkish]{kurimo2010morpho}. The MorphScore data is from \citet{arnett2025why}. The second and third columns ("stem-suffix" and "suffix-suffix") correspond respectively to testing only the one stem-suffix boundary of a word (mimicking how MorphScore works), and testing all boundaries except for that one (for which at least 3 morphemes are needed). The top languages are considered fusional, the bottom agglutinative. Word counts are in \autoref{apx:setup}.}
	\label{tab:seg-results}
\end{table*}

\paragraph{MorphScore.}\label{sec:morphscore}
Micro-averaged precision, recall and $F_1$ are established metrics for measuring morphological boundary recognition \citep[\emph{inter alia}]{kurimo_unsupervised_2006,gronroos_morfessor_2014,bostrom2020byte}.
\citeauthor{arnett2025why} introduce a new metric, MorphScore, with associated datasets.
The MorphScore datasets consist of one morpheme boundary per word, namely the boundary between a stem and its suffix(es).
The MorphScore metric considers micro-averaged recall of these boundaries in two modes: one in which words that appear in both the test set and the tokenizer's vocabulary are left untested, and another in which those words are always counted as correct even if the segmentation does not recall the stem-suffix boundary.
\autoref{tab:seg-examples} shows why tracking only recall, and why only considering the stem-suffix boundary, does not accurately judge morphological alignment.

\begin{table}[H]
	\centering
	\small
	\resizebox{0.7\linewidth}{!}{\begin{tabular}{lll}
    \toprule
     Segmentation & MS & F-$F_1$ \\
     \midrule
     
     \emph{gathered} $\rightarrow$ \stem{gather}\sep\sufone{ed} & 1 & 1.0 \\
     \emph{gathered} $\rightarrow$ \stem{gather}\sufone{e\sep d} & 0 & 0.0 \\
     \emph{gathered} $\rightarrow$ \stem{g\sep a\sep t\sep h\sep e\sep r\sep}\sufone{e\sep d} & 1 & 0.25 \\
     \\
     
     \emph{arabalar\i} $\rightarrow$ \stem{araba}\sep \sufone{lar}\sep \suftwo{\i} & 1 & 1.0 \\
     \emph{arabalar\i} $\rightarrow$ \stem{araba}\sep \sufone{lar}\suftwo{\i} & 1 & 0.5 \\ 
     \emph{arabalar\i} $\rightarrow$ \stem{araba}\sufone{lar}\sep \suftwo{\i} & 0 & 0.5 \\ 
    \bottomrule
\end{tabular}
}
	\caption{Examples of what is being evaluated by full-alignment (F-$F_1$) and MorphScore (MS). Full-alignment refers to evaluating a tokenizer on all morpheme boundaries. MorphScore evaluates the recall of stem-suffix boundaries. Stems are marked \textcolor{highlightgreen}{green}, other colors indicate which characters belong to a refrence morpheme.\vspace{-0.5em}}
	\label{tab:seg-examples}
\end{table}

\noindent Ignoring suffix-suffix boundaries is problematic when relating \als, \fls, and LMs.
Consider a fusional word of the form \textsl{\stem{aaaaa}\sep\sufone{bc}} and an agglutinated word of the form \textsl{\stem{wwwww}\sep\sufone{xx}\sep\suftwo{yy}\sep\sufthree{zz}}.
First, both words have a stem-suffix boundary, but the odds of the stem and suffix sticking together are slightly lower in \als, since the suffix morphs might already form a bigger token as in \textsl{\stem{wwwww}\sep\sufone{xx}\suftwo{yy}\sep\sufthree{zz}}, which \citeauthor{arnett2025why} also observe.
Second, missing a stem-suffix boundary produces highly specific tokens for both, as in \textsl{\stem{aaaaa}\sufone{b\sep c}} or \textsl{\stem{aaaa\sep a}\sufone{bc}}, and \textsl{\stem{wwwww}\sufone{x\sep x}\sep\suftwo{yy}\sep\sufthree{zz}} or \textsl{\stem{wwww\sep w}\sufone{xx}\sep\suftwo{yy}\sep\sufthree{zz}}.
Third, missing an agglutinative suffix-suffix boundary is potentially much worse: in \textsl{\stem{wwwww}\sep\sufone{xx}\suftwo{y\sep y}\sep\sufthree{zz}}, the \textsl{\suftwo{yy}} morph has lost half its length, making it potentially meaningless.
Finally, misses can cascade, as in \textsl{\stem{wwwww}\sep\sufone{xx}\suftwo{y\sep y}\sufthree{z\sep z}} or \textsl{\stem{wwwww}\sep\sufone{xx}\suftwo{y\sep y}\sufthree{zz}}; the three suffix morphemes will be harder for a model to piece together from the embeddings of two tokens.\footnote{This is what happens in Table 3 of \citet{ataman_linguistically_2017}.}
In short: stem-suffix boundaries are not explanatory for \als underperforming to \fls.
Losing one boundary in \als may cause the same performance hit as losing many more boundaries in \fls, making morphological alignment a poor predictor for language modeling.

Using MorphScore, \citeauthor{arnett2025why} find that tokenization for \als is more aligned (higher stem-suffix recall) than for \fls.
This is based on averages across MorphScore's 22 languages, evenly divided between \als and \fls.
Yet, this average hides notable inconsistencies: English (FL) has the second-highest MorphScore of all languages, and five \fls have a higher MorphScore than Turkish (AL).
The conclusion that \als are more aligned than \fls is partially caused by this averaging across groups.
We get back to this grouping issue in \autoref{sec:metrics}.

\paragraph{Full Alignment.} We now use inflectional and derivational segmentations from MorphyNet \citep{batsuren2021morphynet} to measure full alignment. Additionally, we create two  datasets from words with at least three morphemes: one with only the stem-suffix boundary, the other with only the remaining suffix-suffix boundari(es).

The tokenizers \citeauthor{arnett2025why} use for \textbf{H1} are from \citet{chang2024whena}, which are not openly available.
These are monolingual ULM\footnote{\citet{chang2024whena,chang2024goldfisha,arnett2025why} all refer to these as "SentencePiece tokenizers", but SentencePiece itself is a software package \cite{kudo2018sentencepiece}, where the user chooses between ULM \cite{kudo2018subword} or BPE \cite{sennrich2016neural}, resulting in different tokenizers.} tokenizers trained on 10k randomly sampled "lines", with a vocabulary size of 32k.
Instead, we use the monolingual tokenizers from the
\emph{Goldfish} suite of models \citep{chang2024goldfisha}; these also use ULM, but they are trained on more data (1 GiB\footnote{This 1 GiB is "byte-premium-adjusted", see \autoref{sec:h3}.} sampled from a pool of several datasets) and with a vocabulary size of 50k.
\autoref{tab:seg-results} shows the results of full alignment and compares to MorphScore where available.\footnote{We added precision and $F_1$ for MorphScore's boundaries, but it is technically only the \underline{\textbf{Re}} column.}

\paragraph{Findings.} To reliably conclude whether tokenization for \als is more aligned than \fls, and what it implies for LMs, one would ideally use full reference segmentations for a large set of languages and evaluate various tokenization methods.

The data we have limits us to MorphScore, containing 22 languages with about 100 to 2000 examples each, MorphyNet, containing 13 languages\footnote{There are 15 total, but manual inspection showed questionable labels for Italian and Mongolian.} ranging from 100k to over 1 million examples, and MorphoChallenge with about 1000 examples (\autoref{apx:setup}).

The "stem-suffix" and "suffix-suffix" columns in \autoref{tab:seg-results} show that \emph{within} a language, suffix-suffix boundaries are missed much more than stem-suffix boundaries in almost all cases, but at an unpredictable rate.
Thus, only checking alignment for stem-suffix boundaries does not allow assessing the alignment of the boundaries that possibly matter even more (as discussed at the start of this section).

Comparing scores \emph{across} languages shows neither grouping consistently having a higher $F_1$.

\subsection{H2: Tokenization Efficiency}\label{sec:h2}

Whereas alignment refers to morphological correspondence between tokens and morphs, efficiency
refers to whether a tokenizer optimally uses its allocated vocabulary for encoding a text.

\emph{Corpus token count} \citep[\emph{CTC,}][]{schmidt-etal-2024-tokenization} and \emph{Rényi efficiency} \citep[\emph{RE,}][]{zouhar2023tokenization} have been proposed to quantify this. CTC measures how many tokens are needed to encode a given text.
Despite sometimes claimed to measure "compression", it cannot be compared across texts and languages unless (at least) normalized by the length of the source text (yielding the inverse of \emph{mean token length (MTL)}).
RE quantifies the uniformity of the token distribution for a certain text, as measured by its entropy $H_\alpha$ normalized by its maximally achievable entropy $H_0$.
High entropy means the distribution is uniform (flat), low entropy means it is skewed with very frequent and very rare tokens \citep{zipf_human_1949}.
High entropy is desirable since it means the whole vocabulary receives training data, rather than overloading a small number of types.

\als have a smaller affix inventory than \fls because more are used when forming words, so no affix suffers from information scarcity.
\citeauthor{arnett2025why} compute RE and CTC on the multi-parallel FLORES-200 dataset \cite{team2022noa} and indeed find higher RE for tokenizers in \als, although they conclude that this is undesirable.
They find little connection to PPL (surprisingly, since RE, CTC, and PPL were all higher for \als) for the monolingual CLMs by \citet{chang2024whena} in 36 \als and 16 \fls.\footnote{The reported number is 63, the actual 53: \autoref{tab:experimental-variables}.} In \autoref{sec:metrics}, we find that in larger corpora, there are minimal differences between the REs of \fls (e.g.\ Romanian) and \als (e.g.\ Finnish); we argue that CLMs are more affected by the distribution of token bigrams than token unigrams.

\paragraph{Findings.}
To conclude that \als have worse tokenizer efficiency and that this impacts LMs, an ideal experiment would have parallel training and testing data for a large set of languages, and would relate multiple intrinsic metrics (e.g.\ alignment and efficiency) to LM metrics (e.g.\ language characteristics \cite{meister2021languagea} and downstream performance) that are comparable, with no unnecessary grouping. All this is a large-scale effort and outside the scope of the current paper.

The conclusions by \citeauthor{arnett2025why} are arguably not reliable due to comparing monolingual PPLs (\autoref{sec:ppl}) and coarse, unbalanced groupings (\autoref{sec:metrics} and \autoref{tab:experimental-variables}).
Their claim that higher entropy is undesirable also conflicts with \citet{zouhar2023tokenization}.

\subsection{H3: Dataset Size}\label{sec:h3}
More data generally results in better LMs \citep{kaplan_scaling_2020,bousquet_monotone_2022}. When training models on non-parallel monolingual corpora, each model should be supplied with the same amount of information.
Neither the corpus character count (CCC), token count (CTC), or word count (CWC) reliably quantify this due to being confounded by morphology and tokenization.
The \emph{corpus sentence count (CSC)} is less confounded, if sentence boundaries can be found \citep{minixhofer_wheres_2023}.

\paragraph{Byte-premiums.}
To test whether models for \als were just trained on less data,
\citeauthor{arnett2025why} compare PPLs computed for the previously mentioned monolingual \emph{Goldfish} models, whose training data was scaled per language to reflect its
\emph{byte-premium (BP)}. BPs were introduced by \citet{arnett2024bita} to measure how many extra bytes are needed to encode parallel texts in UTF-8 compared to English (due to e.g.\ script or diacritics).
For the Goldfish models, using 10 MiB of English text as reference, a language requiring $3\times$ more bytes to represent got 30 MiB of (non-parallel) training data.
The comparison reported 154 languages,\footnote{The actual number is 149, see \autoref{tab:experimental-variables}.} grouped into 85 \als and 64 \fls.

\begin{table*}
	\centering
	\begin{subtable}[c]{0.75\textwidth}
		\centering
		\small
		\begin{tabular}{llllllp{8em}}
    \toprule
    Experiment & $|$L$|$ & \als & \fls & $|$V$|$ & Tokenizer Data & Metric \\
    \midrule
    \textbf{H1}: Alignment & 22 & 11 & 11 & 32k & 10k lines & MorphScore, PPL$^*$ \\
    \midrule
    \textbf{H2}: Efficiency & 63 (53$^\dag$) & 37$^\ddag$ & 16 & 32k & 10k lines & CTC, RE, PPL$^*$ \\
    \midrule
    \textbf{H3}: Data Size & 154 (149$^\dag$) & 85$^\ddag$ & 64 & 50k & 100 MiB & PPL$^*$ \\
    \bottomrule
\end{tabular}

		\caption{Language-script pairs, morphological groupings, tokenizers, and metrics per hypothesis.}
		\label{tab:experimental-variables}
	\end{subtable}
	\hfill
	\begin{subtable}[c]{0.2\textwidth}
		\centering
		\small
		\begin{tabular}{lr}
    \toprule
    Hypotheses & $|$L$|$ \\
    \midrule
    \textbf{H1} $\cap$ \textbf{H2} & 3 \\
    \textbf{H1} $\cap$ \textbf{H3} & 22 \\
    \textbf{H2} $\cap$ \textbf{H3} & 52 \\
    \textbf{H1} $\cap$ \textbf{H2} $\cap$ \textbf{H3} & 3 \\
    \midrule
    \textbf{H1} $\cup$ \textbf{H2} $\cup$ \textbf{H3} & 145 \\
    \bottomrule
\end{tabular}

		\caption{Language overlap.}
		\label{tab:language-intersection}
	\end{subtable}
	\caption{Experimental conditions in \citet{arnett2025why}. Multiple experimental variables change per hypothesis, making it impossible to know what caused observed effects. $\dag$ means the reported number is incorrect due to null values, which are silently dropped in R. $\ddag$ means the grouping contains languages that are included twice, but written in different scripts. \textbf{H2} has 52 and \textbf{H3} 145 unique languages. All duplicate language-script combinations are ALs. We report \emph{languages}, not language-script combinations in \autoref{tab:language-intersection}. $^*$\Hone \& \Htwo are from \citet{chang2024whena}, \Hthree from \citet{chang2024goldfisha}.
    \vspace{-1em}}
	\label{tab:experiments}
\end{table*}

\citeauthor{arnett2025why} observe that the previously seen PPL gap between \fls and \als shrinks to have a $p$-value of $7.7\%$ when using the Goldfish models, whose datasets controlled for BP.
They conclude from this that there is no longer a performance gap and that BP explains this compared to the original experiments that showed a larger gap (statistical discussion: \autoref{apx:stats}).
Yet, the previous experiments used different vocabulary sizes, training data, and languages (see \autoref{tab:experiments}); all these confounding changes are potential causes for the observed effect.
\begin{mdframed}[
		linewidth=1.5pt,
		topline=false,
		rightline=false,
		bottomline=false,
		leftmargin = 0.1em,
		rightmargin=-0.75em,
		usetwoside=false,
	]
	\small\slshape The results show that after taking into account byte premiums, there is no difference in performance according to morphological typology. [...] This suggests, therefore, that differences that seemed to be driven by morphological typology are actually being driven by disparities in dataset size measurement. -- \citeauthor{arnett2025why}
\end{mdframed}
\paragraph{Findings.}
We agree that BP is an interesting alternative to CSC, but by scaling the datasets using BP, changing the tokenizers, the model sizes, and the languages studied, it becomes impossible determine the effect of BP.
To isolate it, one could do a paired $t$-test between pairs of LMs that share their language, tokenizer, architecture, and test set (preferably part of a fully parallel corpus), with one model trained on e.g.\ 1 $\times$ 10 MiB and the other on $\text{BP}_L$ $\times$ 10 MiB of data.
We would also need a fair metric to assess LM performance across languages (see \autoref{sec:ppl}).
Concluding that BP explains away morphology is not possible otherwise.
A $p$-value of 7.7\% implies that if there truly was no difference between the PPLs of BP-adjusted \als and \fls (the null hypothesis), gaps as high as the one found for the Goldfish models would only occur in 1 out of every 13 repeats of the experiment.
We outline additional issues regarding experimental setups and hypothesis testing in \autoref{apx:hypothesesdiff}.

\subsection{Recap}
In \autoref{tab:experimental-variables}, we list the experimental variables per hypothesis of \citeauthor{arnett2025why}, and in \autoref{tab:language-intersection} which languages they each share.
On top of confounding factors (see \autoref{sec:factors}), only three languages are present in all three hypotheses, meaning conclusions are predominantly drawn about \emph{different} languages, bringing into question their reliability.

\paragraph{Perplexities Across Languages.}\label{sec:ppl}
To answer the central question, a metric to quantify language modeling performance is needed.
Variations\footnote{E.g.\ negative log-likelihood (NLL) or log-perplexity, both monotone transformations and thus interchangeable.} of \emph{perplexity (PPL)} are often used which measure if a model assigns low probabilities to each next token in a test sequence \cite{cotterell2018are,gerz2018language,mielke2019what,park2021morphology,wan2022fairness,chang2024goldfisha}.
As shown, experimental setups commonly use \emph{monolingual} models evaluated on a test set in the model's language.
One model achieving a lower PPL than another would indicate that it is "better".

Comparing PPLs between monolingual models is not straightforward without strong assumptions. Comparing PPLs of different \emph{models} with the same tokenizer and test set \cite{chang2024whena} is valid.
Comparing PPLs of different \emph{segmentations} of the same test text can be compared after rescaling to a shared underlying unit like characters \cite{mielke2019can,bauwens2024bitspercharacter}.
Yet, when comparing PPLs from different models targeting different test sets in different languages, not only does the segmentation change,\footnote{Assuming we are using subword tokenization.} but also the underlying distribution of the test set -- even with parallel texts, see e.g.\ \autoref{tab:ppl}.
The argument made in favor of this comparison is that with parallel texts, models are capturing "semantic information" across languages, which is what we would ideally use.
However, even after transforming PPL into negative log-likelihood, bits-per-character, or relative metrics such as bits-per-English-character, we are still comparing different distributions (texts) using different segmentations.
\begin{table}[ht]
	\centering
	\small
	\resizebox{\linewidth}{!}{\begin{tabular}{llc}
    \toprule
    Model & Sequence & PPL \\
    \midrule
    A & Sabe\toksep jugar\toksep al\toksep ajedrez & 20\\
    B & Do\toksep you\toksep know\toksep how\toksep to\toksep play\toksep chess & 22 \\
    B & Can\toksep you\toksep play\toksep chess & 18 \\
    \bottomrule
\end{tabular}
}
	\caption{Two valid parallel English sentences for a Spanish sentence (same semantic information) with hypothetical PPLs. If we (arbitrarily) select the first parallel English option, suddenly model A is "better" than B, and vice-versa for the second option.}
	\label{tab:ppl}
    \vspace{-1em}
\end{table}
\noindent Grouping monolingual PPLs into morphological categories and performing analyses on them is based on the assumption that these PPLs are drawn from the same distribution.\footnote{Even if we assume they are comparable, we have to deal with outliers.
	Perplexity ranges from 1 to $\infty$, hindering robust, direct comparisons.
	Indeed, the results by \citeauthor{arnett2025why} contain clear outliers (see \autoref{apx:ppl}). The mere existence of these outliers also shows byte-premiums are not the ultimate solution to performance disparities across languages.}
We argue for future research on comparable and informative intrinsic language modeling metrics if we want to find a reliable answer to the central question.

\section{Gradient View of Languages}\label{sec:metrics}
Coarse morphological groupings are useful to talk about general tendencies, not for answering the central question.
Reporting averages over these groupings hides individual language characteristics for both morphological phenomena (fusional--agglutinative scale) and performance differences.

CLMs predict a token after a prior sequence. Intuitively, this is easier if there are fewer valid options to choose from; we can quantify this by measuring per context (the current token) how many possible follow-ups there exist for it in a corpus. Such token \emph{bigram} metrics will be more informative than \emph{unigram} metrics like TTR.\footnote{Imagine a text that enumerates the alphabet. Its TTR is maximal, yet a CLM can deterministically reproduce it.}

\begin{table*}[t]
	\centering
	\tiny
	\resizebox{\linewidth}{!}{\begin{tabular}{ll|rrrr|rrr|rr}
\multicolumn{2}{c|}{} & \multicolumn{4}{c|}{Token Bigrams} & \multicolumn{3}{c|}{Token Unigrams} & \multicolumn{2}{c}{Words} \\
\multicolumn{1}{l}{Language} & \multicolumn{1}{l|}{Grouping$^*$} & \multicolumn{1}{c}{AV} & \multicolumn{1}{c}{$\eta$ ($\downarrow$)} & \multicolumn{1}{c}{AU} & \multicolumn{1}{c|}{LR} & \multicolumn{1}{c}{\scalebox{0.75}{MATTR}} & \multicolumn{1}{c}{MTL} & \multicolumn{1}{c|}{RE} & \multicolumn{1}{c}{$\mathcal S$} & \multicolumn{1}{c}{MWL} \\
\hline
English & Fusional & \gradproxy[2.1211252048993914][4.044892599519379][7.137534477729758]{2.12} & \gradproxy[15.91942205034592][26.33009085190066][40.31395203003415]{15.92} & \gradproxy[50.61045102370512][55.04819832505369][61.53724543067449]{61.08} & \gradproxy[28.950807819582323][43.810328843322246][59.88368570909312]{59.29} & \badproxy[31.77846700951903][37.80174889898552][45.72438686242305]{31.78} & \badproxy[4.695011406724573][5.046404781816551][5.374096019376062]{4.89} & \badproxy[32.262185953590986][35.53231854284905][40.30068772684927]{36.68} & \wordproxy[2.29880945446543][12.11874589210588][16.58109328113736]{9.27} & \wordproxy[5.537347205110425][6.1019656760769765][7.78445446237931]{5.54} \\
French & Fusional & \gradproxy[2.1211252048993914][4.044892599519379][7.137534477729758]{2.39} & \gradproxy[15.91942205034592][26.33009085190066][40.31395203003415]{19.11} & \gradproxy[50.61045102370512][55.04819832505369][61.53724543067449]{57.77} & \gradproxy[28.950807819582323][43.810328843322246][59.88368570909312]{51.55} & \badproxy[31.77846700951903][37.80174889898552][45.72438686242305]{34.27} & \badproxy[4.695011406724573][5.046404781816551][5.374096019376062]{5.08} & \badproxy[32.262185953590986][35.53231854284905][40.30068772684927]{40.30} & \wordproxy[2.29880945446543][12.11874589210588][16.58109328113736]{2.30} & \wordproxy[5.537347205110425][6.1019656760769765][7.78445446237931]{5.91} \\
Dutch & Fusional & \gradproxy[2.1211252048993914][4.044892599519379][7.137534477729758]{3.33} & \gradproxy[15.91942205034592][26.33009085190066][40.31395203003415]{20.75} & \gradproxy[50.61045102370512][55.04819832505369][61.53724543067449]{60.61} & \gradproxy[28.950807819582323][43.810328843322246][59.88368570909312]{43.60} & \badproxy[31.77846700951903][37.80174889898552][45.72438686242305]{33.85} & \badproxy[4.695011406724573][5.046404781816551][5.374096019376062]{5.17} & \badproxy[32.262185953590986][35.53231854284905][40.30068772684927]{37.83} & \wordproxy[2.29880945446543][12.11874589210588][16.58109328113736]{8.36} & \wordproxy[5.537347205110425][6.1019656760769765][7.78445446237931]{6.01} \\
Portuguese & Fusional & \gradproxy[2.1211252048993914][4.044892599519379][7.137534477729758]{3.06} & \gradproxy[15.91942205034592][26.33009085190066][40.31395203003415]{21.31} & \gradproxy[50.61045102370512][55.04819832505369][61.53724543067449]{52.64} & \gradproxy[28.950807819582323][43.810328843322246][59.88368570909312]{51.49} & \badproxy[31.77846700951903][37.80174889898552][45.72438686242305]{35.38} & \badproxy[4.695011406724573][5.046404781816551][5.374096019376062]{4.91} & \badproxy[32.262185953590986][35.53231854284905][40.30068772684927]{36.38} & \wordproxy[2.29880945446543][12.11874589210588][16.58109328113736]{10.64} & \wordproxy[5.537347205110425][6.1019656760769765][7.78445446237931]{5.79} \\
Spanish & Fusional & \gradproxy[2.1211252048993914][4.044892599519379][7.137534477729758]{2.95} & \gradproxy[15.91942205034592][26.33009085190066][40.31395203003415]{22.70} & \gradproxy[50.61045102370512][55.04819832505369][61.53724543067449]{56.97} & \gradproxy[28.950807819582323][43.810328843322246][59.88368570909312]{52.62} & \badproxy[31.77846700951903][37.80174889898552][45.72438686242305]{33.85} & \badproxy[4.695011406724573][5.046404781816551][5.374096019376062]{5.05} & \badproxy[32.262185953590986][35.53231854284905][40.30068772684927]{36.16} & \wordproxy[2.29880945446543][12.11874589210588][16.58109328113736]{9.05} & \wordproxy[5.537347205110425][6.1019656760769765][7.78445446237931]{5.72} \\
Danish & Fusional & \gradproxy[2.1211252048993914][4.044892599519379][7.137534477729758]{3.84} & \gradproxy[15.91942205034592][26.33009085190066][40.31395203003415]{24.12} & \gradproxy[50.61045102370512][55.04819832505369][61.53724543067449]{57.44} & \gradproxy[28.950807819582323][43.810328843322246][59.88368570909312]{38.71} & \badproxy[31.77846700951903][37.80174889898552][45.72438686242305]{33.32} & \badproxy[4.695011406724573][5.046404781816551][5.374096019376062]{4.78} & \badproxy[32.262185953590986][35.53231854284905][40.30068772684927]{35.53} & \wordproxy[2.29880945446543][12.11874589210588][16.58109328113736]{11.91} & \wordproxy[5.537347205110425][6.1019656760769765][7.78445446237931]{5.82} \\
Bulgarian & Fusional & \gradproxy[2.1211252048993914][4.044892599519379][7.137534477729758]{3.37} & \gradproxy[15.91942205034592][26.33009085190066][40.31395203003415]{24.12} & \gradproxy[50.61045102370512][55.04819832505369][61.53724543067449]{52.91} & \gradproxy[28.950807819582323][43.810328843322246][59.88368570909312]{40.74} & \badproxy[31.77846700951903][37.80174889898552][45.72438686242305]{36.37} & \badproxy[4.695011406724573][5.046404781816551][5.374096019376062]{4.86} & \badproxy[32.262185953590986][35.53231854284905][40.30068772684927]{34.88} & \wordproxy[2.29880945446543][12.11874589210588][16.58109328113736]{12.21} & \wordproxy[5.537347205110425][6.1019656760769765][7.78445446237931]{5.97} \\
Swedish & Fusional & \gradproxy[2.1211252048993914][4.044892599519379][7.137534477729758]{3.84} & \gradproxy[15.91942205034592][26.33009085190066][40.31395203003415]{24.18} & \gradproxy[50.61045102370512][55.04819832505369][61.53724543067449]{57.29} & \gradproxy[28.950807819582323][43.810328843322246][59.88368570909312]{35.71} & \badproxy[31.77846700951903][37.80174889898552][45.72438686242305]{35.90} & \badproxy[4.695011406724573][5.046404781816551][5.374096019376062]{5.11} & \badproxy[32.262185953590986][35.53231854284905][40.30068772684927]{39.79} & \wordproxy[2.29880945446543][12.11874589210588][16.58109328113736]{8.73} & \wordproxy[5.537347205110425][6.1019656760769765][7.78445446237931]{6.10} \\
Greek & Fusional & \gradproxy[2.1211252048993914][4.044892599519379][7.137534477729758]{4.20} & \gradproxy[15.91942205034592][26.33009085190066][40.31395203003415]{24.48} & \gradproxy[50.61045102370512][55.04819832505369][61.53724543067449]{51.62} & \gradproxy[28.950807819582323][43.810328843322246][59.88368570909312]{46.81} & \badproxy[31.77846700951903][37.80174889898552][45.72438686242305]{38.71} & \badproxy[4.695011406724573][5.046404781816551][5.374096019376062]{5.11} & \badproxy[32.262185953590986][35.53231854284905][40.30068772684927]{37.44} & \wordproxy[2.29880945446543][12.11874589210588][16.58109328113736]{10.35} & \wordproxy[5.537347205110425][6.1019656760769765][7.78445446237931]{6.15} \\
Romanian & Fusional & \gradproxy[2.1211252048993914][4.044892599519379][7.137534477729758]{3.12} & \gradproxy[15.91942205034592][26.33009085190066][40.31395203003415]{25.09} & \gradproxy[50.61045102370512][55.04819832505369][61.53724543067449]{51.81} & \gradproxy[28.950807819582323][43.810328843322246][59.88368570909312]{51.01} & \badproxy[31.77846700951903][37.80174889898552][45.72438686242305]{37.80} & \badproxy[4.695011406724573][5.046404781816551][5.374096019376062]{5.04} & \badproxy[32.262185953590986][35.53231854284905][40.30068772684927]{36.98} & \wordproxy[2.29880945446543][12.11874589210588][16.58109328113736]{10.52} & \wordproxy[5.537347205110425][6.1019656760769765][7.78445446237931]{5.95} \\
German & Fusional & \gradproxy[2.1211252048993914][4.044892599519379][7.137534477729758]{4.04} & \gradproxy[15.91942205034592][26.33009085190066][40.31395203003415]{26.33} & \gradproxy[50.61045102370512][55.04819832505369][61.53724543067449]{57.29} & \gradproxy[28.950807819582323][43.810328843322246][59.88368570909312]{33.66} & \badproxy[31.77846700951903][37.80174889898552][45.72438686242305]{35.83} & \badproxy[4.695011406724573][5.046404781816551][5.374096019376062]{5.28} & \badproxy[32.262185953590986][35.53231854284905][40.30068772684927]{35.14} & \wordproxy[2.29880945446543][12.11874589210588][16.58109328113736]{12.12} & \wordproxy[5.537347205110425][6.1019656760769765][7.78445446237931]{6.52} \\
Italian & Fusional & \gradproxy[2.1211252048993914][4.044892599519379][7.137534477729758]{3.65} & \gradproxy[15.91942205034592][26.33009085190066][40.31395203003415]{27.10} & \gradproxy[50.61045102370512][55.04819832505369][61.53724543067449]{61.54} & \gradproxy[28.950807819582323][43.810328843322246][59.88368570909312]{59.88} & \badproxy[31.77846700951903][37.80174889898552][45.72438686242305]{37.56} & \badproxy[4.695011406724573][5.046404781816551][5.374096019376062]{5.22} & \badproxy[32.262185953590986][35.53231854284905][40.30068772684927]{38.85} & \wordproxy[2.29880945446543][12.11874589210588][16.58109328113736]{9.39} & \wordproxy[5.537347205110425][6.1019656760769765][7.78445446237931]{6.21} \\
Latvian & Fusional & \gradproxy[2.1211252048993914][4.044892599519379][7.137534477729758]{4.45} & \gradproxy[15.91942205034592][26.33009085190066][40.31395203003415]{28.07} & \gradproxy[50.61045102370512][55.04819832505369][61.53724543067449]{50.99} & \gradproxy[28.950807819582323][43.810328843322246][59.88368570909312]{43.81} & \badproxy[31.77846700951903][37.80174889898552][45.72438686242305]{41.75} & \badproxy[4.695011406724573][5.046404781816551][5.374096019376062]{5.00} & \badproxy[32.262185953590986][35.53231854284905][40.30068772684927]{32.29} & \wordproxy[2.29880945446543][12.11874589210588][16.58109328113736]{15.76} & \wordproxy[5.537347205110425][6.1019656760769765][7.78445446237931]{6.41} \\
Czech & Fusional & \gradproxy[2.1211252048993914][4.044892599519379][7.137534477729758]{4.58} & \gradproxy[15.91942205034592][26.33009085190066][40.31395203003415]{30.07} & \gradproxy[50.61045102370512][55.04819832505369][61.53724543067449]{50.71} & \gradproxy[28.950807819582323][43.810328843322246][59.88368570909312]{41.32} & \badproxy[31.77846700951903][37.80174889898552][45.72438686242305]{43.06} & \badproxy[4.695011406724573][5.046404781816551][5.374096019376062]{4.70} & \badproxy[32.262185953590986][35.53231854284905][40.30068772684927]{35.15} & \wordproxy[2.29880945446543][12.11874589210588][16.58109328113736]{13.67} & \wordproxy[5.537347205110425][6.1019656760769765][7.78445446237931]{6.01} \\
Polish & Fusional & \gradproxy[2.1211252048993914][4.044892599519379][7.137534477729758]{4.74} & \gradproxy[15.91942205034592][26.33009085190066][40.31395203003415]{30.85} & \gradproxy[50.61045102370512][55.04819832505369][61.53724543067449]{50.61} & \gradproxy[28.950807819582323][43.810328843322246][59.88368570909312]{43.80} & \badproxy[31.77846700951903][37.80174889898552][45.72438686242305]{44.51} & \badproxy[4.695011406724573][5.046404781816551][5.374096019376062]{5.25} & \badproxy[32.262185953590986][35.53231854284905][40.30068772684927]{35.76} & \wordproxy[2.29880945446543][12.11874589210588][16.58109328113736]{12.75} & \wordproxy[5.537347205110425][6.1019656760769765][7.78445446237931]{6.68} \\
Slovak & Fusional & \gradproxy[2.1211252048993914][4.044892599519379][7.137534477729758]{4.70} & \gradproxy[15.91942205034592][26.33009085190066][40.31395203003415]{31.12} & \gradproxy[50.61045102370512][55.04819832505369][61.53724543067449]{51.43} & \gradproxy[28.950807819582323][43.810328843322246][59.88368570909312]{44.68} & \badproxy[31.77846700951903][37.80174889898552][45.72438686242305]{43.04} & \badproxy[4.695011406724573][5.046404781816551][5.374096019376062]{4.82} & \badproxy[32.262185953590986][35.53231854284905][40.30068772684927]{34.91} & \wordproxy[2.29880945446543][12.11874589210588][16.58109328113736]{13.39} & \wordproxy[5.537347205110425][6.1019656760769765][7.78445446237931]{6.13} \\
Slovenian & Fusional & \gradproxy[2.1211252048993914][4.044892599519379][7.137534477729758]{4.09} & \gradproxy[15.91942205034592][26.33009085190066][40.31395203003415]{32.04} & \gradproxy[50.61045102370512][55.04819832505369][61.53724543067449]{52.85} & \gradproxy[28.950807819582323][43.810328843322246][59.88368570909312]{48.35} & \badproxy[31.77846700951903][37.80174889898552][45.72438686242305]{40.42} & \badproxy[4.695011406724573][5.046404781816551][5.374096019376062]{4.77} & \badproxy[32.262185953590986][35.53231854284905][40.30068772684927]{33.74} & \wordproxy[2.29880945446543][12.11874589210588][16.58109328113736]{13.66} & \wordproxy[5.537347205110425][6.1019656760769765][7.78445446237931]{5.88} \\
Lithuanian & Fusional & \gradproxy[2.1211252048993914][4.044892599519379][7.137534477729758]{6.26} & \gradproxy[15.91942205034592][26.33009085190066][40.31395203003415]{33.62} & \gradproxy[50.61045102370512][55.04819832505369][61.53724543067449]{52.82} & \gradproxy[28.950807819582323][43.810328843322246][59.88368570909312]{44.35} & \badproxy[31.77846700951903][37.80174889898552][45.72438686242305]{44.11} & \badproxy[4.695011406724573][5.046404781816551][5.374096019376062]{5.00} & \badproxy[32.262185953590986][35.53231854284905][40.30068772684927]{32.26} & \wordproxy[2.29880945446543][12.11874589210588][16.58109328113736]{16.58} & \wordproxy[5.537347205110425][6.1019656760769765][7.78445446237931]{6.61} \\
Finnish & Agglutinative & \gradproxy[2.1211252048993914][4.044892599519379][7.137534477729758]{7.14} & \gradproxy[15.91942205034592][26.33009085190066][40.31395203003415]{36.83} & \gradproxy[50.61045102370512][55.04819832505369][61.53724543067449]{55.05} & \gradproxy[28.950807819582323][43.810328843322246][59.88368570909312]{28.95} & \badproxy[31.77846700951903][37.80174889898552][45.72438686242305]{45.72} & \badproxy[4.695011406724573][5.046404781816551][5.374096019376062]{5.37} & \badproxy[32.262185953590986][35.53231854284905][40.30068772684927]{34.60} & \wordproxy[2.29880945446543][12.11874589210588][16.58109328113736]{16.23} & \wordproxy[5.537347205110425][6.1019656760769765][7.78445446237931]{7.78} \\
Hungarian & Agglutinative & \gradproxy[2.1211252048993914][4.044892599519379][7.137534477729758]{6.69} & \gradproxy[15.91942205034592][26.33009085190066][40.31395203003415]{39.11} & \gradproxy[50.61045102370512][55.04819832505369][61.53724543067449]{56.24} & \gradproxy[28.950807819582323][43.810328843322246][59.88368570909312]{31.37} & \badproxy[31.77846700951903][37.80174889898552][45.72438686242305]{41.73} & \badproxy[4.695011406724573][5.046404781816551][5.374096019376062]{5.05} & \badproxy[32.262185953590986][35.53231854284905][40.30068772684927]{34.10} & \wordproxy[2.29880945446543][12.11874589210588][16.58109328113736]{14.63} & \wordproxy[5.537347205110425][6.1019656760769765][7.78445446237931]{6.78} \\
Estonian & Agglutinative & \gradproxy[2.1211252048993914][4.044892599519379][7.137534477729758]{6.27} & \gradproxy[15.91942205034592][26.33009085190066][40.31395203003415]{40.31} & \gradproxy[50.61045102370512][55.04819832505369][61.53724543067449]{55.89} & \gradproxy[28.950807819582323][43.810328843322246][59.88368570909312]{34.39} & \badproxy[31.77846700951903][37.80174889898552][45.72438686242305]{43.66} & \badproxy[4.695011406724573][5.046404781816551][5.374096019376062]{5.22} & \badproxy[32.262185953590986][35.53231854284905][40.30068772684927]{34.58} & \wordproxy[2.29880945446543][12.11874589210588][16.58109328113736]{14.87} & \wordproxy[5.537347205110425][6.1019656760769765][7.78445446237931]{6.96} \\
\end{tabular}
}
	\caption{We propose to use gradient proxies of morphology that operate on token \emph{bigrams} (\autoref{fig:av}) within "words" (pretokens): the variety of a type's accessors (AV), their uniqueness (AU), and the Shannon efficiency of their distribution ($\eta$). We report averages over types in the tokenizer's vocabulary that appear at least once and were not filtered (see \autoref{apx:filter}); the fraction of types excluded from each average is its lexicalization ratio (LR). We also give existing metrics operating on token \emph{unigrams}: moving-average type-token-ratio (MATTR), micro-average characters per token (mean token length; MTL), and Rényi efficiency (RE). Last are word-based metrics: tokens per character averaged per word ($\mathcal S$) and mean word length (MWL). All metrics are calculated on EuroParl \cite{koehn2005europarla} using the same tokenizers as \autoref{tab:seg-results}. $^*$Groupings taken from \citet{arnett2025why}. The gradient in the columns ranges from its minimum to maximum and are intended to highlight how the metrics differ. We sort by $\eta$. For AU and LR, the top three are highlighted yellow, the bottom three orange. For visual clarity, all metrics except for AV, MTL, and MWL are multiplied by 100.}  %
	\label{tab:europarl}
    \vspace{-1em}
\end{table*}

\paragraph{Accessor Variety.}
\citet{harris_phoneme_1955} first suggested to count the variety of predecessor and successor units of a given string, where unusual spikes would imply the string's edges delineated something meaningful like a morpheme or word.
\citet{feng_accessor_2004} coined \emph{accessor variety (AV)} as the minimum of predecessor and successor variety.
\citet{wu_finding_2018} applied this to subword tokens to learn BPE merges. We use ULM tokens.

Formally, let $V$ be a subword vocabulary, $t_1,t_2 \in V$, and $f(t_1,t_2)$ be the amount of times a token of type $t_1$ is followed immediately by a token of type $t_2$ in a corpus. The sets
\begin{equation}
	\begin{aligned}
		\A_L(t) & = \{t'\in V \mid f(t',t) > 0\} \\
		\A_R(t) & = \{t'\in V \mid f(t,t') > 0\}
	\end{aligned}
\end{equation}
are respectively the \emph{left accessors} (predecessors) and \emph{right accessors} (successors) of $t \in V$. We similarly define a \emph{left AV} and \emph{right AV}:
\begin{equation}
	\AV_L(t) = |\A_L(t)| \qquad \AV_R(t) = |\A_R(t)|.
\end{equation}
Since AV is bounded by the \emph{total accessors (TA)}
\begin{equation}
	\begin{aligned}
		\TA_L(t) & = \sum_{t'\in \A_L(t)} f(t',t)  \\
		\TA_R(t) & = \sum_{t'\in \A_R(t)} f(t,t'),
	\end{aligned}
\end{equation}
it can be confined to a fixed range, which we denote as \emph{accessor uniqueness (AU)}:
\begin{equation}
	\AU_L(t) = \frac{\AV_L(t)}{\TA_L(t)} \quad\; \AU_R(t) = \frac{\AV_R(t)}{\TA_R(t)}.
\end{equation}
AU is analogous to TTR, except for token \emph{bigrams}.
TTR has been criticized for its dependency on corpus size, which can be relieved by computing it in fixed-size windows and averaging across those \citep{covington_cutting_2010}. Thus, for $\AV(t)$ and $\AU(t)$, each $t$ keeps a 1000-accessor window.

Since each type has a \emph{distribution} of accessors on its left and right, we measure their \emph{Shannon efficiency}: how close they are to a uniform distribution. For the right accessors, this is
\begin{equation}
	\scalemath{1.0}{
		\eta_R(t) = \frac{1}{\hat H_0^R}
		\sum_{t'\in\A_R(t)} \frac{f(t,t')}{\TA_R(t)} \log_2 \frac{f(t,t')}{\TA_R(t)}}
\end{equation}
where $\hat H_0^R(t) = \log_2\min\{|\dom\A_L|,\TA_R(t)\}$ is the maximally achievable entropy with $\TA_R(t)$ samples taken from the $|\dom\A_L| \leq |V|$ right accessors that appear in the corpus. \autoref{fig:av} shows a diagram of the above metrics. In what follows, we filter out types with little to no accessors (see \autoref{apx:filter}).

Finally, there are two ways of applying the bigram metrics: either to characterize \emph{morphological complexity} or \emph{data difficulty}.
For the former, we look at "intra-word" tokens, meaning we pretokenize our input and subsequently tokenize the pretokens.
This is what is \citet{feng_accessor_2004} do.
Alternatively, we can forego the pretokenization step and calculate AV directly.
The former gives us an idea about the token-to-token ambiguity within pretokens (closer to morphology), the latter gives an idea of the token-to-token ambiguity without directly relating to words or pretokens (closer to the data).
See \autoref{apx:filter} for more details.

\paragraph{Multi-Parallel Results.}
In \autoref{tab:europarl}, we calculate our metrics on a multi-parallel aligned subset of the EuroParl \cite{koehn2005europarla} corpus.
The alignment is for the sake of removing confounds of data sizes and domains, not to draw conclusions based on the parallel meaning (c.f. \autoref{sec:ppl}).
In the next section, we loosen this restriction and expand our language set.

\autoref{tab:europarl} shows that AV recovers the coarse groupings, with \als having the highest AV. Additionally, within \fls, a more fine-grained view of morphological complexity is revealed. For instance, higher AV values point to compounding languages (e.g.\ German and Danish) as opposed to the lower ones (e.g.\ English and Romanian).
The shape of the accessor distribution as summarized by $\eta$ follows the same trend, being higher (more uniform) for \als.
These results for AV and $\eta$ are both crucial in light of our hypothesis above, i.e.\ that the difficulty of causal language modeling, and hence the source of higher PPL, is having \emph{more} and \emph{more equally likely follow-up options} at each token.
This is what AV and $\eta$ measure.
Therefore, if the hypothesis is correct, then higher AV and $\eta$ are causally linked to higher PPL, thus explaining the \als and \fls gap.

The word-based metrics recover the groupings somewhat, but are less directly related to CLMs, unless they also use words instead of subword tokens.
Additionally, we have to define reference \emph{words} which is another potential confound.
The token unigram metrics look rather even across the languages in EuroParl, showing less correspondence with the other metrics.
Since these estimators become more accurate with more data, their low variance calls into question higher-variance results computed for much smaller corpora like FLORES-200.\footnote{EuroParl has 211k multi-parallel lines with Italian as a pivot, \mbox{FLORES-200} has about 2k when combining the dev and test splits, which are not available for all languages; full multi-parallel alignment results in about 1k lines.\label{foot:data-sizes}}

Lastly, AV operates on \emph{tokens}, which means its applicable to other units.
For character- or byte-level tokenizers, we can still get an estimate of the degree of choice of accessors for a given type.

\begin{figure*}[ht]
	\centering
	\includegraphics[width=\linewidth]{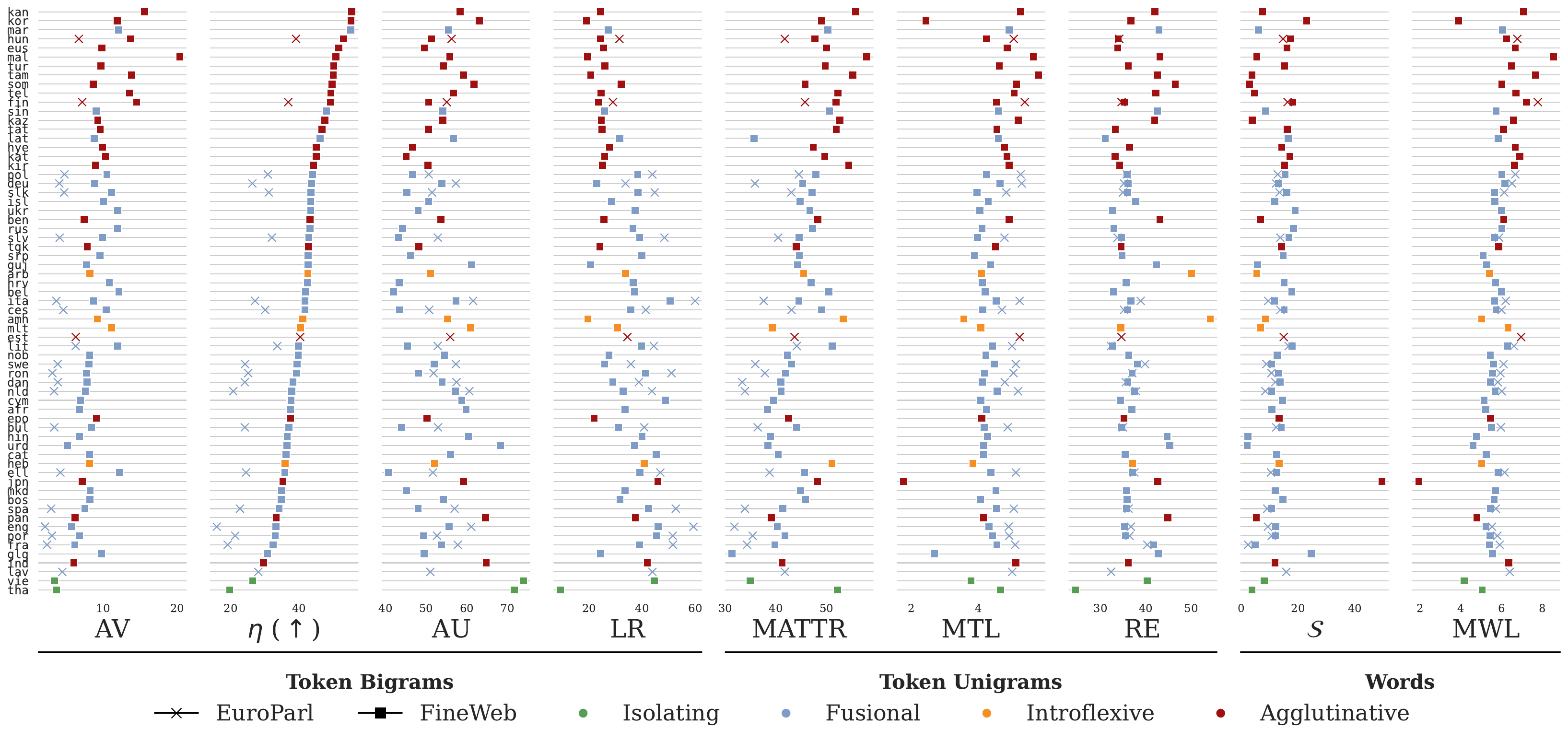}
	\caption{Metrics across EuroParl and FineWeb. The bigram metrics are calculated within pretokens, \autoref{fig:additional-sorting-results} contains results without pretokens. Similar to \autoref{tab:europarl}, we sort by $\eta$ to show how (dis)similar the metrics are from existing metrics. EuroParl (EP) contains 21 languages; FineWeb (FW) 63. FW $\cap$ EP $=$ 19; FW $\cup$ EP $=$ 65. Full results in table form are in \autoref{apx:results}. The added coarse groupings are \emph{Isolating} languages, which tend to have little to no inflection and use few morphemes per word, and \emph{Introflective} (or non-concatinative) languages, which modify roots and tend to use little to no morphemes.}
	\label{fig:results}
    \vspace{-1em}
\end{figure*}

\paragraph{Expanded Results.}
When we loosen the restriction of multi-parallelism, we can expand our language coverage.
We use data from \mbox{FineWeb 1 \& 2} \cite{penedo2024fineweb,penedo2025fineweb2}.
Our selection is based on (1) the language has to have a 1 GiB Goldfish tokenizer and (2) it has to have 200k lines of available data.
All languages thus have similar \emph{amounts} of data for both datasets.
\autoref{fig:results} shows the combined results.

Starting with AV and $\eta$, we see a clear domain influence, with EuroParl always having lower values compared to FineWeb.
The opposite is true for MATTR, suggesting that EuroParl is more lexically dense than web texts, but in a repeating fashion.
The expanded language coverage shows even more clearly that coarse groupings hide information.
While the top languages are agglutinative, and the bottom ones isolating, the middle shows the need for a gradient proxy.
Perhaps unsurprisingly, MWL is another decent proxy for morphological complexity, but as mentioned, LMs do not tend to use words.
Language modeling difficulty is tokenizer-dependent.
Similarly, token unigram metrics measure noticeably different phenomena, and, as discussed in \autoref{sec:h2}, these are less important for relating morphology to language modeling compared to the bigram metrics.

\paragraph{Morphology or Data.}
We have shown results \emph{with} pretokenization given the link with morphology (requiring "words").
Results \emph{without} (\autoref{tab:av-pretok}) also show \als near the top, but especially languages using long words written in systems with large character inventories (e.g. abugidas).

This relates to \textbf{H3} by \citet{arnett2025why}: some writing systems require more UTF-8 bytes to encode and more data mitigates this.
However, characteristics of written language are \emph{part of} writing systems.
Or, as put by \citet{gorman2023myths} \emph{``A writing system is, at its base, a linguistic analysis of the language it is used to write."}
\citeauthor{arnett2025why}'s conclusion could be explained in opposite terms: since morphology is encoded in writing systems, we need to account for it (byte-premiums).

\section{Conclusion}
We identify confounding factors to consider in order to reliably answer the question of \emph{whether, and how, morphology relates to language modeling}. These factors imply "ideal" experiments, from which to work backwards to what is feasible.

We re-evaluate three hypotheses from \citet{arnett2025why} for why there might be a causal language modeling performance gap between agglutinative and fusional languages: morphological alignment of tokenization, tokenization efficiency, and dataset size.
We show recall of stem-suffix boundaries (MorphScore) is not full alignment and outline how alignment relates to LMs (\autoref{sec:h1}).
We agree and re-confirm that token unigram metrics are poor explanations for the gap (\autoref{sec:h2}).
We disagree with the conclusion that dataset size explains away modeling difficulty caused by morphology and suggest methodological improvements (\autoref{sec:h3}). %

Finally, we introduce token bigram metrics (accessor variety and entropic efficiency) that quantify the ambiguity language models face, and show they are gradient proxies of morphology, providing a new hypothesis for why causal language models might struggle more with agglutination.

\newpage
\section*{Limitations}\label{sec:limitations}
\paragraph{Additional Factors.}
The confounding factors we discuss especially relate to the question of \emph{whether, and how, morphology relates to language modeling}.
We acknowledge there are many more confounding factors, such as architecture choices, domain, data quality, translation effects of parallel data, among others.
We do not discuss these since, while all important, they should ideally stay fixed when trying to answer the central question.

\paragraph{Segmentation Availability.}
Our full alignment analysis from \autoref{sec:h1} relies on high quality reference segmentations.
These are rare and their language coverage is quite limited, which prevents us from making broad conclusions.

\paragraph{Text Features and Morphology.}
Our metrics from \autoref{sec:metrics} are \emph{not} metrics for morphological complexity as understood in the linguistics literature (E- and I-complexity, see \autoref{sec:related}).
Instead, they are proxies for morphological phenomena as seen \emph{through the lens of} a particular tokenizer (here, monolingual ULM tokenizers) over a particular corpus (here, EuroParl or FineWeb). The assumption for this to work -- but empirically, this seems to be correct -- is that patterns in how tokenizers construct words mimic patterns of how words are constructed from morphological systems.

\paragraph{Dataset Size.} We do not compute our metrics on the corpus for which \citeauthor{arnett2025why} have PPL values, i.e.\ FLORES-200 \citep{team2022noa}, since we found it too small to get stable results, see \autoref{foot:data-sizes}. EuroParl is significantly larger and showed more stable metrics, which is what we use as our multi-parallel corpus. For our non-parallel corpus, but using a consistent number of \emph{lines}, we use FineWeb. We make sure the number of lines between EuroParl and FineWeb are comparable ($\sim$200k).

\paragraph{Morphological Groupings.}
There are many ways to characterize morphological systems of languages.
The course groupings are commonly used, but have also long been criticized, such as by \citet{sapir1921language}, who notes: \textit{``In any case it is very difficult to assign all known languages to one or other of these groups, the more so as they are not mutually exclusive."}
Our use of these groupings is purely for the sake of comparison with previous work.

\vfill
\newpage
\ifx\review\undefined
	\section*{Acknowledgments}
We thank Kushal Tatariya for suggestions, comments, and proofreading an earlier draft of this work.
We also thank Catherine Arnett for answering our questions and making the analysis scripts openly available.
Finally, we thank the anonymous reviewers for their suggestions.
WP and TB are funded by a KU Leuven Bijzonder Onderzoeksfonds C1 project with reference C14/23/096.
The computational resources and services used were provided by the VSC (Flemish Supercomputer Center), funded by the Research Foundation - Flanders (FWO) and the Flemish Government - department EWI.

\fi

\bibliographystyle{acl_natbib}
\bibliography{bib/bibliography}

\cleardoublepage
\appendix

\vfill
\newpage
\section{Pretokenization and Filtering}\label{apx:filter}
\subsection{Pretokenization}
For the word-based approaches discussed in \autoref{sec:metrics}, we preprocess sentences by splitting them on spaces and punctuation.
For languages where this pretokenization step is problematic, Japanese and Thai in our case, we use a dedicated word segmenter \cite{mccann2020fugashi,phatthiyaphaibun2023pythainlp}.

Tokenization for our bigram metrics happens within pretokens, and as in \citet{feng_accessor_2004}, tokens of neighboring pretokens \emph{cannot see each other}. Instead, the first token sees a "dummy" token to its left that always counts as a unique accessor no matter how many times it has been seen. The same is true for the last token.\footnote{The idea is that there is such high flexibility in what lies beyond a word boundary that one can work with the upper limit that there is always a different accessor there. This higher flexibility is confirmed in \autoref{tab:av-pretok}.}

Unlike \citet{feng_accessor_2004}, we do not include these dummy accessors in $\A_L(t)$ and $\A_R(t)$. Instead, we count them separately as $b_L(t)$ and $b_R(t)$, resp.\ the amount of times a type $t$ occurs as the first and last token of a word. Note that $f(t) = \TA_L(t) + b_L(t) = \TA_R(t) + b_R(t)$. The fraction of dummy accessors is the \emph{boundary ratio (BR)}:
\begin{equation}
	\scalemath{0.85}{
		\BR_L(t) = \frac{b_L}{\TA_L(t)+b_L} \quad    \BR_R(t) = \frac{b_R}{\TA_R(t)+b_R}
	}.
\end{equation}

Lastly, we also include results of our bigram metrics without pretokenization in \autoref{apx:results}.

\subsection{Filtering}
Per language, after tokenizing the dataset and counting accessors, we retroactively apply two filtering steps to the vocabulary (modifying the counts appropriately) with the goal of reducing noise in the statistics computed from them.

First, we filter out all types that contain at least one character whose value for the Unicode character property \emph{general category}\footnote{\scriptsize\href{https://www.unicode.org/versions/Unicode17.0.0/core-spec/chapter-4/\#G124142}{unicode.org/versions/Unicode17.0.0/core-spec/chapter-4/\#G124142}} is either \textsl{Punctuation} or \textsl{Digit}. In short: we filter out all types matching the regular expression
\begin{verbatim}
    .*(\p{Punct}|\p{Digit}).*
\end{verbatim}
The resulting vocabulary $V'$ is assumed to come entirely from the language's lexicon for that corpus.

Since we are interested in the distribution of tokens (i.e.\ non-boundaries) around each type (the tokenization equivalent of morphology), we further exclude all types which are almost never accessed by other types and thus mostly by dummies, i.e.\
\begin{equation}
	\bcancel V = \{t\in V' \mid \min\{\BR_L(t),\BR_R(t)\} \geq 0.95\}.
\end{equation}
These types could be said to be \emph{lexicalized} by the tokenizer and have such sparse or empty accessor distributions that including summary metrics for those distributions would merely be noise.

We then call the fraction of types excluded from the vocabulary its \emph{lexicalization ratio (LR)}:
\begin{equation}
	\LR = \frac{|\bcancel V|}{|V'|}.
\end{equation}

\section{Statistical Sidenotes}\label{apx:stats}
We want to address some of the consequences of particular decisions made by \citet{arnett2025why} in their statistical analyses. Specifically:

\begin{itemize}
    \item Designing their study as a difference of hypothesis tests rather than a hypothesis test of a difference (\autoref{apx:hypothesesdiff});
    \item Defining the null hypothesis and insignificance as the success of a treatment, meaning the $p$-value does not express whether the result is uncommon in the control;
    \item Using different statistical tests to prove the existence of the gap between \fls and \als versus proving its disappearance (\autoref{apx:hypothesistools});
    \item Lacking Bonferroni correction, raising the chances of encountering at least one significant hypothesis test (\autoref{apx:bonferroni});
    \item Assuming PPL has no outliers or is not normally distributed, either way invalidating hypothesis tests and correlations (\autoref{apx:ppl});
    \item Duplicating measurements for CTC and RE, arbitrarily lowering $p$-value (\autoref{apx:duplicates});
    \item Large, skewed predictor distribution in regression, causing highly significant but highly unpredictive regression coefficients (\autoref{apx:largesample});
    \item Suggesting that causation implies correlation (\autoref{apx:causality}).
\end{itemize}

\subsection{Effect of designing the experiments as a \emph{difference of hypothesis tests}, rather than a \emph{hypothesis test of differences}}\label{apx:hypothesesdiff}
In essence, what \citeauthor{arnett2025why} study is the effect of a treatment on a group of subjects.

\subsubsection{Hypothesis test of differences}
\begin{figure}
    \centering

    \begin{tikzpicture}[>=stealth, every node/.style={font=\small}]
        \def\firstTick{1.5}
        \def\secondTick{2.15}
        \def\thirdTick{3}
        
        \draw[->] (-1.5,0) -- (4,0);
        
        \draw (0,0.15) -- (0,-0.15);
        \node[below=2pt] at (0,-0.15) {0};
        
        \draw (\firstTick,0.15) -- (\firstTick,-0.15);
        \node[above=1pt] at (\firstTick,0.15) {$\Delta_{\alpha}$};
        
        \draw (\secondTick,0.15) -- (\secondTick,-0.15);
        \node[above=1pt] at (\secondTick,0.15) {\color{gray}$\Delta_{\text{after}}$};

        \draw (\thirdTick,0.15) -- (\thirdTick,-0.15);
        \node[above=1pt] at (\thirdTick,0.15) {$\Delta_\text{before}$};

        \draw[->,thick] (\firstTick,0.0) -- ($(\firstTick,0.0)-(0.3,0.0)$);
    \end{tikzpicture}
    \caption{One-sided hypothesis test for a significant reduction in an initially positive difference. Everything left of $\Delta_\alpha$ is significant.}
    \label{fig:deltahyp}
\end{figure}
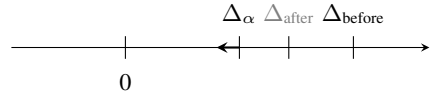

Conventionally, such studies are laid out according to the following recipe:
\begin{enumerate}
    \item Compute a statistic $S$ on the group before the treatment: $s_\text{before}$.
    \item Apply the treatment.\footnote{Applying the treatment to a second group (the first being a control) is also possible, but less practical here.}
    \item Compute the same statistic $S$ on the group after the treatment: $s_\text{after}$.
    \item Assuming the treatment has no effect ($H_0$), formulate a hypothesis test for the statistic such that the more effect the treatment has, the more unlikely $s_\text{after}$ would appear if characterizing the population by $s_\text{before}$. Compute the $p$-value, i.e.\ the probability of $S$ taking on all values even more unlikely than $s_\text{after}$. 
    \item Conclude that the treatment has significant effect ($H_1$) if $p$ is lower than a prespecified threshold $\alpha$. The probability that this conclusion is incorrect, is $p$.
\end{enumerate}
To study what causes the disparity in PPL between \fls and \als, this template might look like:
\begin{itemize}
    \item The group of subjects are a set of languages.
    \item The statistic $S$ is the \emph{gap in average PPL}, $\bar X_1 - \bar X_2$, between the \fls and \als.
    \item The treatment is different for \Hone, \Htwo, and \Hthree. Each of them tries to "explain the gap", which should be designed as a treatment that could make the gap disappear. For each language:
    \begin{itemize}
        \item \Hone: train 1 model with a more and 1 with a less morphologically aligned tokenizer;
        \item \Htwo: train 1 model with a more and 1 with a less compressive tokenizer;
        \item \Hthree: train 1 model with and 1 without byte-premium-scaled training data;
    \end{itemize}
    here, "more" and "less" can be measured continuously using respectively alignment $F_1$ (or MorphScore), CTC (or Rényi efficiency), and byte-premiums.
    
    \item The hypothesis test would measure whether the \emph{gap has decreased significantly}. This calls for a \emph{one-sided} hypothesis test: assigning \fls and \als to the subscripts 1 and 2 such that $\Delta_\text{before} = \bar X_{1,\text{before}} - \bar X_{2,\text{before}} > 0$, a significant treatment would make the difference $\Delta_\text{after} = \bar X_{1,\text{after}} - \bar X_{2,\text{after}}$ move significantly far away to the left of $\Delta_\text{before}$ past some value $\Delta_\alpha < \Delta_\text{before}$, as shown in \autoref{fig:deltahyp}. (That is: either the gap becomes smaller in absolute value, or the sign flips and the absolute value can be anything.)
\end{itemize}
Because it is assumed $\bar X_{i,\text{before}}$ and $\bar X_{i,\text{after}}$ are normally distributed, so are $\Delta_\text{before}$ and $\Delta_\text{after}$ and thus $Y = \Delta_\text{before} - \Delta_\text{after}$ as well, with the sum of the variances of the four means as its variance:
\begin{equation}
    S_Y^2 = \frac{S_{1,\text{before}}^2}{n_{1,\text{before}}} + \frac{S_{2,\text{before}}^2}{n_{2,\text{before}}} + \frac{S_{1,\text{after}}^2}{n_{1,\text{after}}} + \frac{S_{2,\text{after}}^2}{n_{2,\text{after}}}.
\end{equation}
This means $Y$ satisfies the conditions described by \citet{welch_generalization_1947} for $Y/S_Y \sim t(\nu)$ (with $\nu$ given by \citet[Eq.\ 28]{welch_generalization_1947}, a Welch-Satterthwaite equation). Therefore, the hypothesis test
\begin{equation}
    H_1 \,\text{ if }\, \frac{\Delta_\text{before} - \Delta_\text{after}}{S_Y} 
    > t_{1-\alpha,\nu}
\end{equation}
works. That is, when the expected gap between the average PPL for \fls and \als after a treatment is \emph{not} below the expected gap before, the measured gap $\Delta_\text{after}$ will fall below
\begin{equation}
    \Delta_\alpha = \Delta_\text{before} - t_{1-\alpha,\nu} \,S_Y
\end{equation}
only with probability $\alpha$. In short, if we use the fact that $\Delta_\text{after} < \Delta_\alpha$ as the decision rule to decide that $\E[\Delta_\text{after}] < \E[\Delta_\text{before}]$, we are incorrect with probability $\alpha$ when actually $\E[\Delta_\text{after}] \geq \E[\Delta_\text{before}]$.

The effectiveness of the treatment can additionally be assessed by just comparing $\bar X_{1,\text{before}}$ to $\bar X_{1,\text{after}}$ using a usual Welch $t$-test, or comparing $\bar X_{2,\text{before}}$ to $\bar X_{2,\text{after}}$. However, this does not say anything about what the treatment does to the gap \emph{between} the averages; it could be that both averages drop significantly after treatment, but by the same amount, and hence the gap does not change.

\subsubsection{Difference of hypothesis tests}
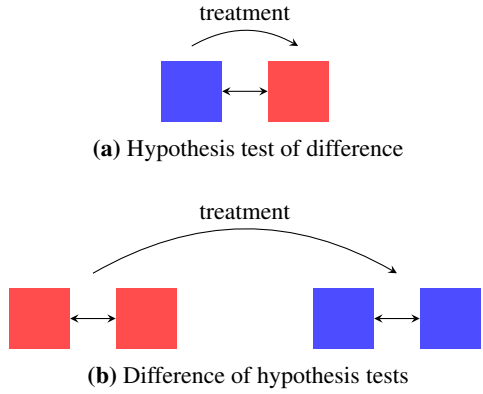
\begin{figure}[ht]
    \centering
    \begin{subfigure}{\linewidth}
        \centering
        \begin{tikzpicture}[>=stealth, every node/.style={font=\small}]
    \def\s{0.8}       %
    \def\innergap{0.6} %
    
    \fill[blue!70] (0,0) rectangle (\s,\s);
    \fill[red!70] (\s+\innergap,0) rectangle (2*\s+\innergap,\s);
    
    \draw[<->] (\s,0.5*\s) -- (\s+\innergap,0.5*\s);
    
    \draw[->, bend left=30] (\s/2,\s+0.2) to node[above]{treatment} (\s+\innergap+\s/2,\s+0.2);
\end{tikzpicture}
        \caption{Hypothesis test of difference}
    \end{subfigure}
    \begin{subfigure}{\linewidth}
        \centering
        \vspace*{1em}
        \begin{tikzpicture}[>=stealth, every node/.style={font=\small}]
    \def\s{0.8}      %
    \def\innergap{0.6} %
    \def\outergap{4} %
    
    \fill[red!70] (0,0) rectangle (\s,\s);
    \fill[red!70] (\s+\innergap,0) rectangle (2*\s+\innergap,\s);
    
    \draw[<->] (\s,0.5*\s) -- (\s+\innergap,0.5*\s);
    
    \fill[blue!70] (\outergap,0) rectangle (\outergap+\s,\s);
    \fill[blue!70] (\outergap+\s+\innergap,0) rectangle (\outergap+2*\s+\innergap,\s);
    
    \draw[<->] (\outergap+\s,0.5*\s) -- (\outergap+\s+\innergap,0.5*\s);
    
    \draw[->, bend left=30] (\s+0.5*\innergap,1.0) to node[above]{treatment} (\outergap+\s+0.5*\innergap,1.0);    
\end{tikzpicture}
        \caption{Difference of hypothesis tests}
    \end{subfigure}
    \caption{The two experimental setups discussed in \autoref{apx:hypothesesdiff}. Each box is a statistic. Each double arrow is a hypothesis test. The red boxes are the values discussed in the text to be desired as causing a significant hypothesis test (i.e.\ rejecting $H_0$) when the treatment \emph{is} effective.}
    \label{fig:treatments}
\end{figure}

\citet{arnett2025why} do not follow the above template. In particular, rather than comparing difference statistics as per above, they compare averages; since there are four averages, they do \emph{two} hypothesis tests for determining if a treatment is significant: one\footnote{Note that this test is not actually reported in the paper.} comparing $\bar X_{1,\text{before}}$ to $\bar X_{2,\text{before}}$, and another comparing $\bar X_{1,\text{after}}$ to $\bar X_{2,\text{after}}$. A treatment is deemed significant if it causes unequal decisions between the two tests, and in particular, because the first test is significant, a treatment is deemed significant if the second test is \emph{not} significant. \autoref{fig:treatments} symbolically represents the difference with the above test.

Equivalently, these two tests compare $\Delta_\text{before}$ and $\Delta_\text{after}$ to 0 rather than comparing them to each other, respectively testing
\begin{equation}
    H_1 \,\text{ if }\, \left|\frac{\Delta_\text{before}-0}{S_{\Delta_\text{before}}}\right|  > t_{1-\alpha/2,\nu}
\end{equation}
and
\begin{equation}
    H_1 \,\text{ if }\, \left|\frac{\Delta_\text{after}-0}{S_{\Delta_\text{after}}}\right| > t_{1-\alpha/2,\nu}.
\end{equation}
This system of hypothesis tests is inappropriate for several reasons. Firstly, despite the sign of the gap being known to the tester, the second test is two-sided: therefore, if a treatment is so good at closing the gap between \fls and \als that in fact \als become \emph{easier} to model than \fls, this test will conclude that the treatment "cannot explain the gap" if the effect is large enough (because the absolute gap will be bigger). Secondly, these tests do not consider the size of the decrease in the gap, but simply the size of the gap itself. That means that if a treatment causes a tiny decrease in the gap yet pushes it just past the $p=\alpha$ threshold, this treatment will be said to "explain the gap" despite its small effect.

Lastly, in both tests, the hypotheses are inverted: the gap between \fls and \als is assumed to \emph{not exist} in both cases when doing the $t$-test, despite the point of the preliminary analysis being that there \emph{is} a gap. For the second test, because the \emph{absence} of a gap is the desired result of the treatments, this means that the null hypothesis is that the treatment \emph{works}. While this would be a straightforward way to set up a hypothesis test comparing averages (if one chooses to use it) and while there is no rule saying the null hypothesis should be the status quo, extreme care should be taken interpreting $p$-values and the role of the significance level $\alpha$: the latter is no longer the false-positive rate of the treatment (the chosen, guaranteed, small probability that a treatment is detected as reducing the gap, given the ground truth that the gap stays) but the false-negative rate (the chosen, guaranteed, small probability that a treatment is detected as not reducing the gap, given the ground truth that the gap is gone). Only one of these rates can be controlled (by setting $\alpha$). Since the status quo is that there is a gap, the assumption should be that this is the ground truth. Therefore, it should also be assumed problematic to fix the probability conditioned on the opposite of this ground truth.

Normally, the lower a $p$-value, the more confidence we have that we are not mistaken in thinking the treatment works. With every decreasing order of magnitude, a $p$-value gives higher confidence; in principle, the $p$ value can keep approximating $0$ indefinitely. When swapping hypotheses, however, a $p$-value \emph{higher} than $\alpha$ is seen as giving confidence in the treatment, but it is unknown to what degree. What a confidence of $p=5\%$ or $p=10\%$ or $p=90\%$ tells about the treatment is unclear.

This particular setup also allows interpreting the conclusion to always favor the treatment. Normally, decreasing $\alpha$ makes it harder for $p$-values of treatments to be considered significant. When swapping hypotheses, this stricter $\alpha$ (5\%, 1\%, 0.001\%, ...) causes almost all $p$-values to accept the null hypothesis (the treatment working).

Let us now consider the $p$-value that is found, and what is concluded about it:
\vfill
\pagebreak
\begin{linequote}
    The Goldfish models exhibit numerically higher perplexity for agglutinative ($M=143.62$) than fusional languages ($M=132.63$), but this difference is not statistically significant ($t(137.36)=1.180$, $p=0.077$). Therefore, after taking byte premiums into effect, the Goldfish models \underline{do not exhibit the same }p\underline{erformance }g\underline{a}p that was demonstrated in previous research and in Section 3 above.\\-- \citeauthor{arnett2025why}
\end{linequote}

The meaning of this 7.7\% is that \emph{if} there is truly no gap, averages even further apart than this would rarely occur -- 7.7\% of all re-executions of the experiment, or 1 in every 13. Yet, this rarity is used to \emph{reject} that there is a gap. This is correct according to the hypothesis test as formulated, but is not a strong case.

\begin{figure*}[t]
    \centering
    \includegraphics[width=0.9\linewidth]{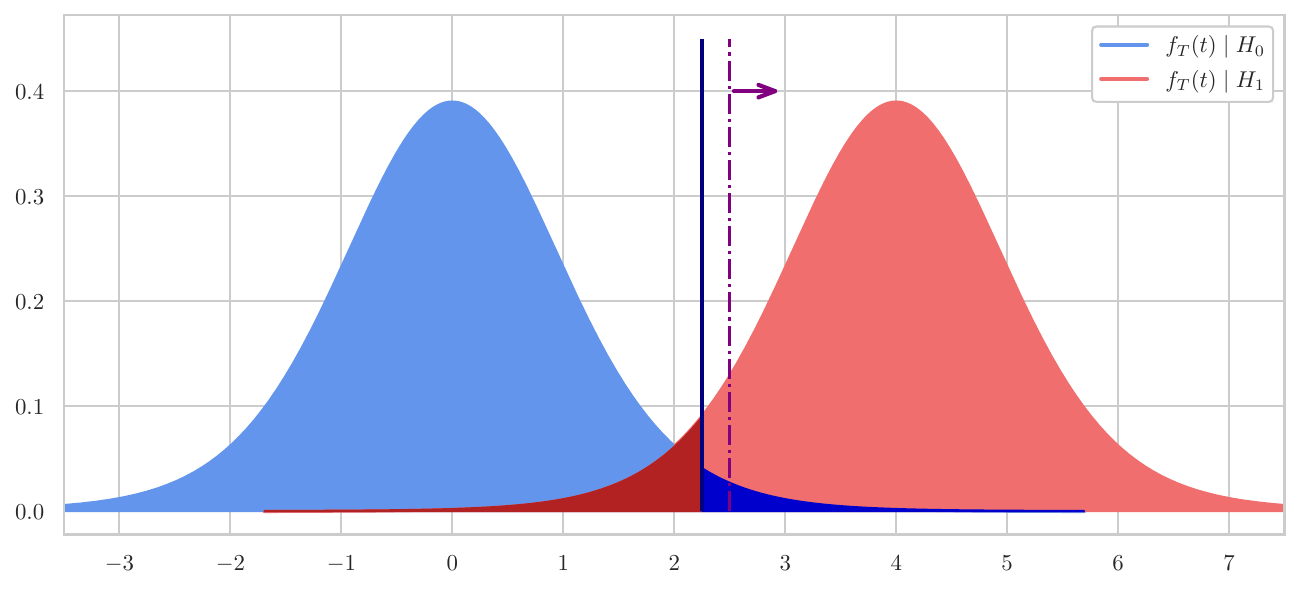}
    \caption{Distribution of a $T$-test statistic under the null hypothesis (e.g.\ "no gap; treatment worked") and the alternative hypothesis (e.g.\ "gap; treatment did not work"). The purple line indicates the hypothesis threshold $t_{\alpha,\nu}$ under $H_0$. The blue line indicates a value of $T$ which is not significant (it is to the left of $t_\alpha$, or equivalently, its $p$-value under the null hypothesis -- the dark blue area -- is bigger than $\alpha$), yet it would be more likely under $H_1$ (the dark red area is bigger than the dark blue area) despite causing $H_1$ to be rejected.}
    \label{fig:alternative}
\end{figure*}

It is in fact impossible to conclude, based on this $p$-value, that "the Goldfish models do not exhibit the same performance gap that was demonstrated". Consider the situation where the ground truth -- which is unknown -- is that there \emph{is} a gap. In that case, the $t$-test statistic is drawn from an unknown distribution, and it could be that it has a \emph{higher $p$-value} (i.e. it is less out-of-place) in this ground truth distribution. One possible scenario where this happens is sketched in \autoref{fig:alternative}. Here, the fact that $p$ was bigger than $\alpha$ is irrelevant, as it is still smaller than the $p$-value of the alternative hypothesis (the existence of a gap, i.e.\ the treatment not working).

\subsection{Effect of using hypothesis tests of different type to measure before and after}\label{apx:hypothesistools}
Across \Hone, \Htwo, and \Hthree, the only treatment that is deemed significant is \Hthree. This is concluded using the inverted $t$-test discussed above. However, this $t$-test comparing mean PPLs of \fls and \als is never actually reported for the "before" case, despite the data being available to do so. Instead, several hypothesis tests for regression coefficients predicting PPL from morphological type are given. From what we could gather, the pre-analysis by \citeauthor{arnett2025why} (Section 3) is intended to serve as stand-ins for the missing $t$-test.  %

In the quote above, it was stated that 
\begin{linequote}
    (...) the Goldfish models do not exhibit the same performance gap that was demonstrated (...) in Section 3 above.
\end{linequote}
Yet, section 3 reads:
\begin{linequote}
    This section describes three analyses that show lower performance for agglutinative languages.\\
    (...) there is still a significant effect of morphological type, where agglutinative languages had higher perplexities than fusional languages.\\
    (...) there is still a significant effect of morphological type, where fusional languages show better performance than agglutinative languages.\\
    (...) morphological type still explains additional variance ($\chi^2(3) = 3.3324$, $p=0.02$).\\
    (...) we found a robust performance gap between agglutinative languages and fusional languages.
\end{linequote}
All that is quantified is how predictive morphological type is in a linear regression with PPL as response, but this has no bearing on the size of the gap. Presumably the statements comparing \fls to \als are about their average PPLs, but no size nor significance is stated.

The lack of a baseline $t$-test analogous to the one in \Hthree also means that when $\alpha$ was chosen for the latter, there was no precedent set from a previous test, so it could have been set at the time the $p$-value was generated, as mentioned above.

\subsection{Effect of many hypothesis tests}\label{apx:bonferroni}
All main hypotheses (i.e.\ the three treatments \Hone, \Htwo, \Hthree) are studied with multiple significance tests (8, 4, and 2, respectively). If it so happens that the ground truth is that no alternative hypotheses hold, then the probability of finding a significant result in a set of $m$ tests is
\begin{equation}\begin{aligned}
    P(\text{any test sig.}\mid H_0) &= \sum_{i=1}^m P(\text{test $i$ sig}\mid H_0) \\
    &= \alpha + \hdots + \alpha \\
    &= \alpha\cdot m
\end{aligned}\end{equation}
assuming each test uses the same significance level $\alpha$. Thus, the odds of seeing rare (significant) events across the study increases proportional to the amount $m$ of tests done, known as the \emph{multiple comparisons problem}. A simple way to mitigate this is to adjust the significance level using a \mbox{(Holm-)Bonferroni} correction \cite{holm1979simple}. In its most basic form, it replaces $\alpha$ by $\alpha' = \alpha/m$.

Note that although $p$-values and $\alpha$ guarantees are concerned with repeated computation of a specific statistic from a specific distribution, the Bonferroni correction should be applied even if each individual test is only performed once, by the above equation. (Nevertheless, some tests by \citeauthor{arnett2025why} are partially about overlapping data, meaning their test decisions are not independent.)

\subsection{Effect of large values on mean and $t$-test}\label{apx:ppl}
\begin{figure}[ht]
	\centering
	\hspace*{-0.5em}\includegraphics[width=1.1\linewidth]{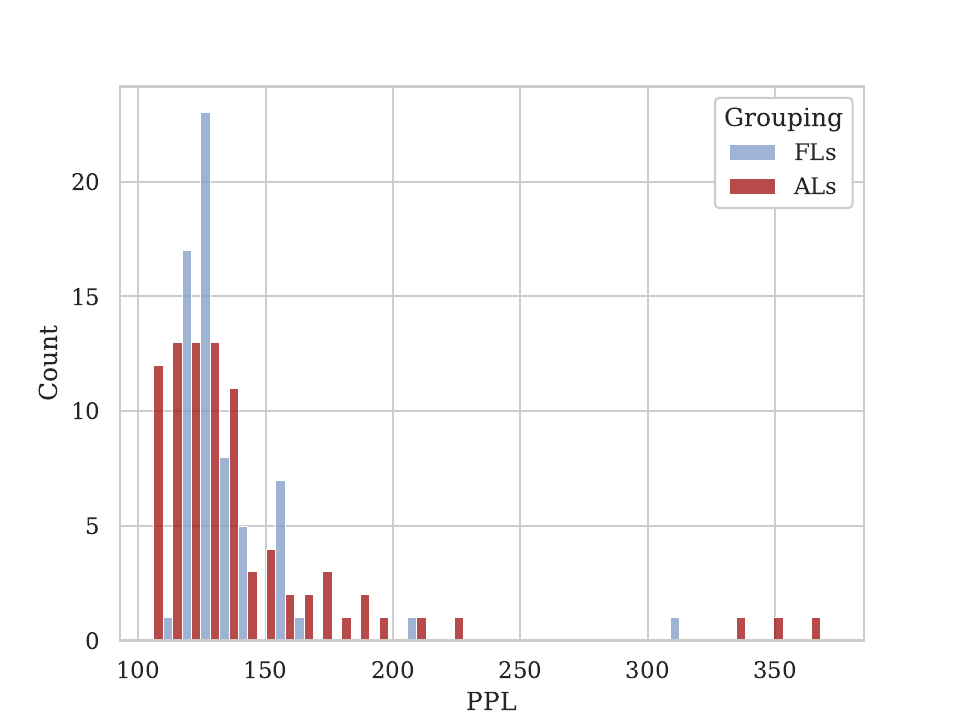}
	\caption{Distribution of PPL values for the languages used in \textbf{H3} in \citet{arnett2025why}.}
	\label{fig:ppl}
\end{figure}

In \Hthree, the means of the PPL distributions of \fls and \als are computed and compared using a Welch $t$-test. In \autoref{fig:ppl}, we see that although most of the models trained by \citeauthor{arnett2025why} achieve a PPL between 110 and 140, there are several much larger values, namely four that lie above 300. Note also that one morphological grouping has 1 such value (\fls) and the other has 3 (\als).

If we assume it unproblematic to compare and aggregate PPL values across languages (despite our objections in \autoref{sec:ppl}). Even then, the existence of such much larger values is always problematic for the statistical tools used: either the tools break down due to the outlying values, or the tools are inapplicable due to a distribution mismatch.

\subsubsection{If large values are outliers}
If the large values are considered outliers of an otherwise normally distributed PPL, then they should be filtered out to not distort downstream statistics.

The \emph{robustness} of a statistical estimator is the ability of its value to withstand perturbation due to outliers. \emph{Breakdown} happens when the value can be changed arbitrarily much by replacing some amount of samples in the dataset by outliers \citep{rousseeuw_robust_2011}. In particular, the most basic sum-based estimators break down with just 1 outlying sample: the sample \emph{mean} and corrected sample \emph{(co)variance} are respectively
\begin{align}
    \bar X &= \frac{1}{n}\sum_{i=1}^n X_i \label{eq:mean}\\
    S_{XY} &= \frac{1}{n-1}\sum_{i=1}^n (X_i - \bar X) (Y_i - \bar Y) \\
    S_X^2  &= S_{XX} = \frac{1}{n-1}\sum_{i=1}^n (X_i - \bar X)^2 \label{eq:variance}
\end{align}
from which other tools for drawing conclusions are derived, such as \emph{correlation} and the \emph{hypothesis $t$-test statistic}
\begin{align}
    R_{XY} &= \frac{S_{XY}}{\,\sqrt{S_{XX}\,S_{XY}}\,} \\[0.5em]
    T &= \frac{\bar X_1 - \bar X_2}{S_T} \label{eq:t-test}
\end{align}
with various definitions of $S_T$ all based on $S_X^2$.

\paragraph{Example.} A list of samples $x = [1,2,3,4,k]$ has median $\tilde x = 3$ for any $k$ between 3 and $+\infty$. Meanwhile, the mean $\bar x = (10+k)/5$ is 3 for $k = 5$, 30 for $k = 140$ and 300 for $k = 1490$. Its correlation with a list of samples $y = [10,20,30,40,50]$ is $r_{xy} = 1.00$ when $k = 5$, but drops to $r_{xy} = 0.89$ when $k = 10$, drops to $r_{xy} = 0.80$ when $k = 20$, and drops to $r_{xy} = 0.72$ when $k = 100$.

\citeauthor{arnett2025why} compare \emph{means} using a \emph{Welch $t$-test} for the PPLs in \Hthree, and compute \emph{correlation} between PPL and CTC in \Htwo. All three methods are not robust, meaning the results are distorted due to the outlying PPLs.

\subsubsection{If large values are expected}
If the large values are not considered outliers, then the assumption is that the distribution from which PPL values are drawn is one with a large right tail, and thus not a normal distribution. The consequence is that both the $p$-value and the acceptance of the null hypothesis, resulting from the Welch $t$-test, are invalidated.

When doing a two-sided $t$-test using the statistic in \autoref{eq:t-test}, its value is compared to the quantiles $t_{\alpha/2,\nu}$ and $t_{1-\alpha/2,\nu}$ of Student's $t$-distribution with $\nu$ degrees of freedom. The statistic is considered significant at significance level $\alpha$ (e.g.\ $1\%$) if it is even less than the former or even more than the latter. That is when the null hypothesis is rejected. The meaning of "at significance level $\alpha$" is that under the null hypothesis, if the experiment were repeated many times on the same population the original sample set came from, the \emph{fraction of experiments} that would turn out significant is $\alpha$. In other words, the probability that the test statistic $T$ falls outside $[t_{\alpha/2,\nu}\, ;\, t_{1-\alpha/2,\nu}]$ is $\alpha$. Since we reject the null hypothesis when this occurs, put differently, $\alpha$ is the probability of a type-I error. The only reason we can be certain of this probability is that the quantile $t_{\alpha/2,\nu}$ is exactly the value for $q$ for which
\begin{equation}\label{eq:probaq}
    P(T \leq q) = \alpha/2
\end{equation}
holds, because $T \sim t(\nu)$ and therefore we know
\begin{equation}\label{eq:cdf}
    P(T \leq q) = \text{CDF}_{t(\nu)}(q)
\end{equation}
and only then can \autoref{eq:cdf} inform \autoref{eq:probaq} to get
\begin{equation}\begin{aligned}
    \text{CDF}_{t(\nu)}(q) &= \alpha/2 \\
    q &= \text{CDF}^{-1}_{t(\nu)}(\alpha/2) \\
      &\equiv t_{\alpha/2,\,\nu}
\end{aligned}\end{equation}
When $\nexists \mu,\sigma : X_i \sim \Normal(\mu,\sigma^2)$, we know \autoref{eq:mean} is no longer normally distributed, which, since it is used in \autoref{eq:t-test}, means $\nexists \nu: T \sim t(\nu)$, and thus \autoref{eq:cdf} no longer holds. That means we do not know for which $q$ \autoref{eq:probaq} holds. Conversely, using $q = t_{\alpha/2,\nu}$, we do not know what significance level we are working with, so we do not know what the probability for a type-I error is.

\begin{figure*}[ht]
    \centering
    \begin{subfigure}{0.49\linewidth}
        \centering
        \includegraphics[width=\linewidth]{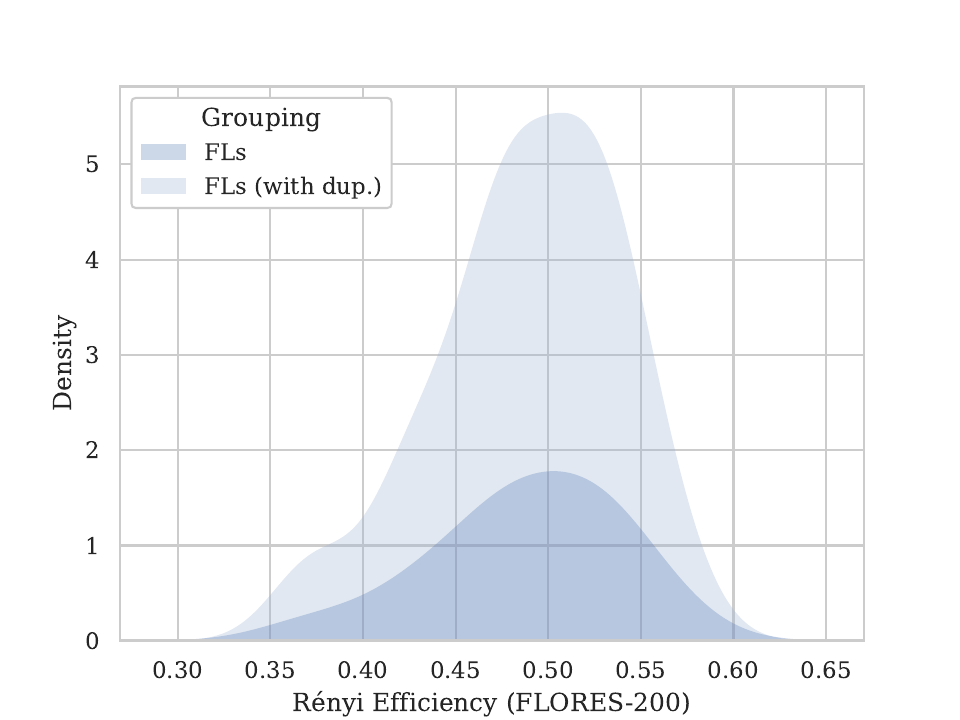}
    \end{subfigure}
    \hfill
    \begin{subfigure}{0.49\linewidth}
        \centering
        \includegraphics[width=\linewidth]{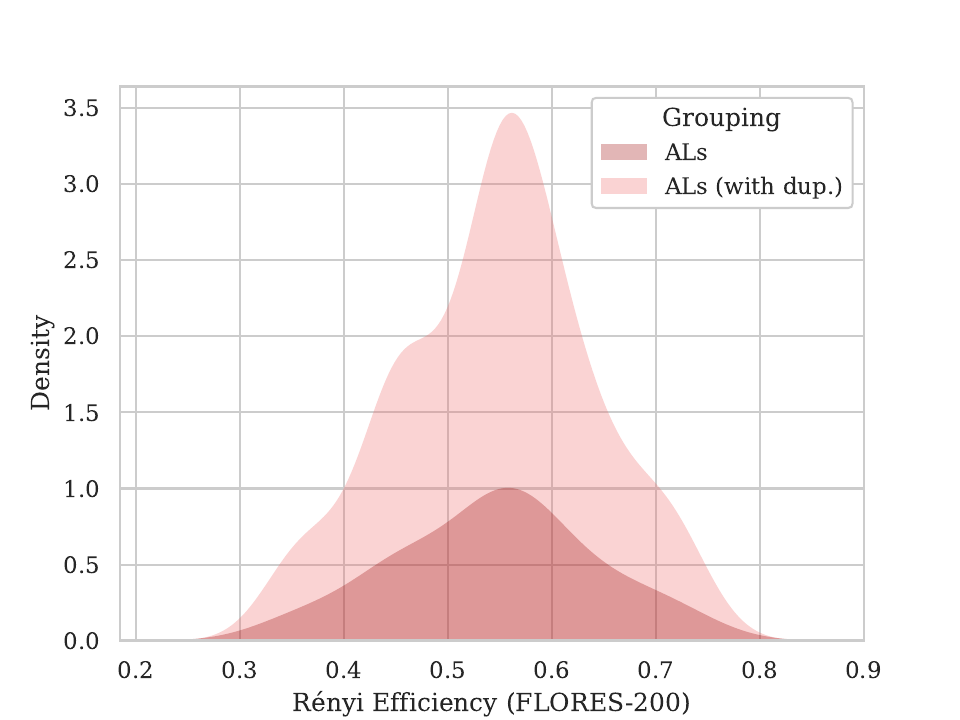}
    \end{subfigure}
    \caption{Sample distributions used for \textbf{H2} with and without deduplication. The density scale is such that the \emph{sum} of the areas under both distributions is 1, while keeping each area proportional to the amount of samples it describes.}
    \label{fig:duplicates}
\end{figure*}

\begin{figure*}
    \includegraphics[width=0.975\linewidth]{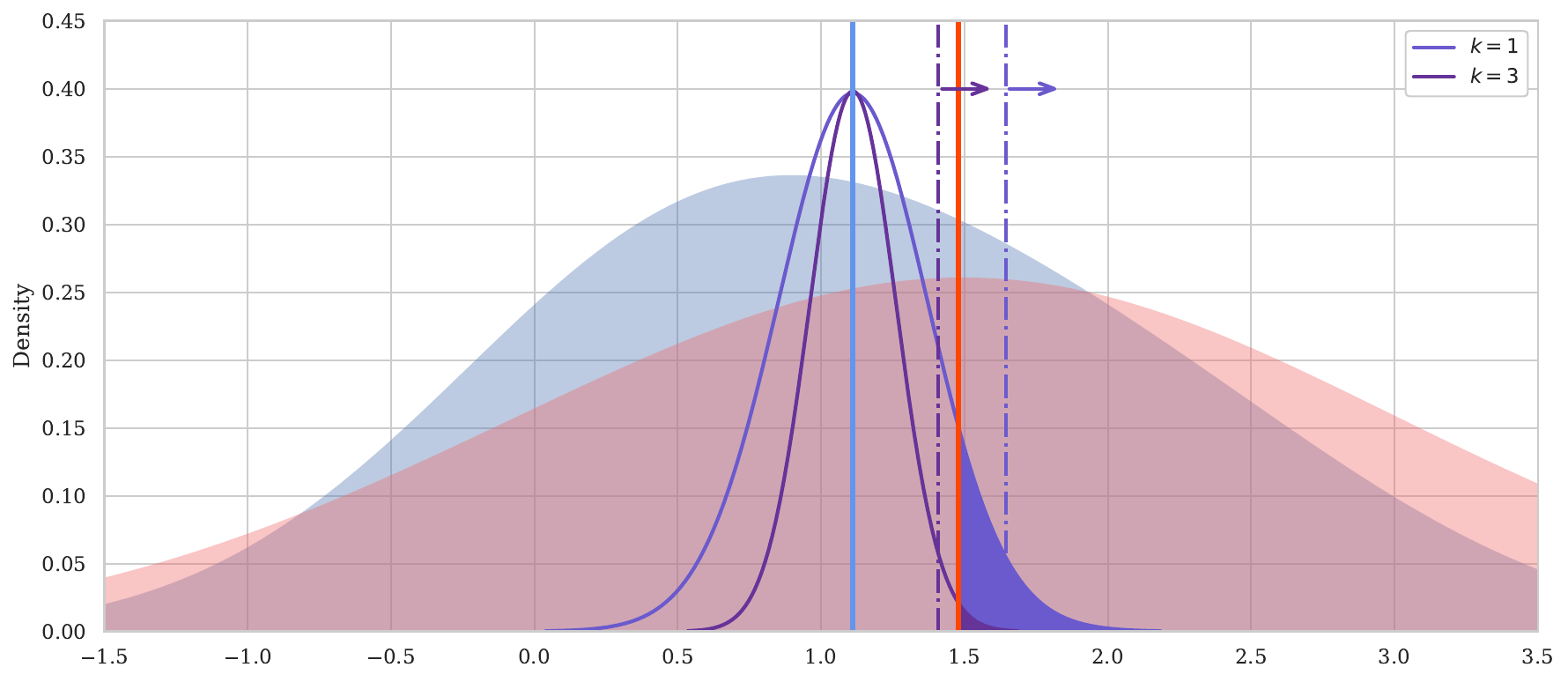}
    \caption{Example of a Welch $t$-test between two samples for which the conclusion changes after including samples $3\times$.}
    \label{fig:hypotheses}
\end{figure*}

\subsubsection{Conclusion}
As a first step for comparing PPLs, either the PPL samples should be made normally distributed through some transform, and/or the outliers should be filtered out (e.g.\ using robust statistics) to draw reliable conclusions from the data.

\subsection{Effect of duplicating data in $t$-test}\label{apx:duplicates}
In \Htwo, a Welch $t$-test is done to compare the mean CTC of tokenizers for \fls ($n_1$) versus \als ($n_2$), and the same thing is done for the mean Rényi efficiency. A significant difference is found for the latter and a nearly significant difference for the former. However, when looking into the R scripts accompanying the paper, we found that before performing the $t$-tests, each of the $n_1$ and $n_2$ measurements is included $3\times$, i.e.\ each measurement is represented as three measurements rather than one.
We show this in \autoref{fig:duplicates}, which incudes deduplicated measurements by language codes. The effect of this is an artificially lowered $p$-value in the hypothesis test, as we show below.

\subsubsection{Visual intuition}
Given two samples from two different normal distribution, a Welch $t$-test asks whether the mean of one is located significantly further from the other to conclude that they are not the same. For a one-sided test for whether the second distribution is to the right of the first, this conclusion is drawn when
\begin{equation}\begin{aligned}
    H_1 \text{ if } T = \frac{\bar X_2 - \bar X_1}{S_u} &> t_{1-\alpha,\nu} \\
                 \bar X_2 &> \bar X_1 + t_{1-\alpha,\nu} \cdot S_u
\end{aligned}
\end{equation}
Visually, on a horizontal axis representing values of $X$, we could mark three points to represent the test: $\bar x_1$, $\bar x_2$, and a threshold which lies $t_{1-\alpha,\nu} \cdot s_u$ to the right of $\bar x_1$. If $\bar x_2$ falls to the right of this threshold, then it is significantly different from $\bar x_1$.

\paragraph{Example.} We take two samples from two normally distributed populations with unequal variances, $\Normal(1,1^2)$ and $\Normal(1.4,1.05^2)$, each of size $n_1 = 25$ and $n_2 = 25$.

In \autoref{fig:hypotheses}, we draw a solid blue line for $\bar x_1$ and a solid red line for $\bar x_2$. The rest is in light purple: the non-standard $t$-distribution $f(t) = \bar x_1 + s_u\cdot t$ is drawn around $\bar x_1$, the $\alpha = 2.5\%$ threshold is a dot-dash line at $t_{1-\alpha,\nu} \cdot s_u$ from the center, the $p$-value of $\bar x_2$ is colored as the area under the null-hypothesis distribution past $\bar x_2$. Visually, the hypothesis test can be executed either by checking whether the red line is to the right of the dot-dash line ($t > t_{1-\alpha,\nu}$), or whether the colored area is smaller than 2.5\% of the total area ($p < \alpha$) under the drawn curve.

We see that the light purple area is large and that the red line stays to the left of the light purple dot-dash line. Thus, we \emph{cannot} conclude from the data that $\bar x_2$ is so significantly far from $\bar x_1$ that its sample was probably taken from a different population.

On the same figure, we repeat the above drawing procedure but after concatenating the samples to themselves twice, resulting in $n_{1,*}=n_{2,*}=75$ measurements for the both samples. We use dark purple this time.

Now we see a much smaller purple area and we see that the red line is now \emph{right of} the dot-dash line.\footnote{We will see below that the dot-dash line has moved left for two reasons: $s_{u,*} < s_u$ and $t_{1-\alpha,\nu_*} < t_{1-\alpha,\nu}$.} The hypothesis test now results in the conclusion that $\bar x_2$ \emph{is} significantly far from $\bar x_1$ supported by a small $p$-value below $\alpha$, despite not having made any more measurements.

\subsubsection{Mathematical proof}
This hypothesis test is of the form
\begin{equation}
    H_0 \text{ if } T = \frac{\bar X_1 - \bar X_2}{S_u} \in [t_{\alpha/2,\nu};t_{1-\alpha/2,\nu}].
\end{equation}
Using the shorthand $V_i = \frac{S_i^2}{n_i}$ with $S_i^2$ given by \autoref{eq:variance}, the unpooled variance estimator $S_u^2$ is
\begin{equation}
    S_u^2 = V_1 + V_2
\end{equation}
and $\nu$ is given by a Welch–Satterthwaite equation
\begin{equation}
    \nu \approx \frac{
        \displaystyle(V_1+V_2)^{2}
    }{ 
        \displaystyle \,
            \frac{V_1^2}{n_1-1} + \frac{V_2^2}{n_2-1}
    \,}
\end{equation}

Now consider two samples of $n_1$ and $n_2$ measurements. When duplicating each measurement by a factor $k$, the following happens to the above quantities (where "$*$" denotes the new situation):
\begin{equation}
    \bar X_* = \frac{1}{n_*} \sum_{j=1}^{n_*} X_j = \frac{1}{kn} \sum_{i=1}^{n} k\,X_i 
    = \bar X
\end{equation}
and thus
\begin{equation}
\begin{aligned}
    S^2_* &= \frac{1}{n_*-1}\sum_{j=1}^{n_*} (X_{j,*} - \bar X_*)^2 \\
    &= \frac{1}{kn-1}\sum_{i=1}^{n} k (X_i - \bar X)^2 \\
    &= \frac{1}{n-1/k}\sum_{i=1}^{n} (X_i - \bar X)^2 = \frac{n-1}{n-1/k}S^2 \\
    &\approx S^2 \\
\end{aligned}
\end{equation}
and thus
\begin{equation}
    V_* = \frac{S_*^2}{n_*} = \frac{S^2_*}{kn} \approx \frac{S^2}{kn} = \frac{1}{k}V
\end{equation}
and thus
\begin{equation}
    S_{u,*}^2 = V_{1,*} + V_{2,*} \approx \frac{1}{k}(V_1 + V_2) = \frac{1}{k} S_u^2
\end{equation}
and thus
\begin{equation}
    T_* = \frac{\bar X_{1,*} - \bar X_{2,*}}{S_{u,*}} = \frac{\bar X_{1} - \bar X_{2}}{S_{u}/\sqrt k} = \sqrt k \cdot T.
\end{equation}
Also,
\begin{equation}\begin{aligned}
    \nu_* &\approx \frac{
        \displaystyle(V_{1,*}+V_{2,*})^{2}
    }{ 
        \displaystyle \,
            \frac{V_{1,*}^2}{n_{1,*}-1} + \frac{V_{2,*}^2}{n_{2,*}-1}
    \,} \\ 
    &\approx \frac{
        \displaystyle \left(\frac{V_{1}}{k}+\frac{V_{2}}{k}\right)^{\!2}
    }{ 
        \displaystyle \,
            \frac{\displaystyle\left(\frac{V_{1}}{k}\right)^{\!2}}{kn_{1}-1} + \frac{\displaystyle\left(\frac{V_{2}}{k}\right)^{\!2}}{kn_{2}-1}
    \,} \\
    &= \frac{
        \displaystyle \frac{1}{k^2}(V_1+V_2)^2
    }{ 
        \displaystyle \,
            \frac{1}{k^3}\left(\frac{V_1^2}{n_1 - 1/k} + \frac{V_2^2}{n_2 - 1/k}\right)
    \,} \\
    &\approx k\cdot\nu
\end{aligned}\end{equation}
and thus 
\begin{equation}
(t_{1-\alpha/2,\,\nu})_* = t_{1-\alpha/2,\,\nu_*} = t_{1-\alpha/2,\,k\nu}\;.
\end{equation}
Therefore, we can conclude the following: in the original dataset, for a hypothesis test to turn out significant, the $T$ statistic (proportional to the gap between the means) was required to be so large that
\begin{equation}
    H_1 \text{ if } |T| > t_{1-\alpha/2,\,\nu}
\end{equation}
whereas in the inflated dataset, a significant hypothesis can already reached when
\begin{equation}\begin{aligned}
    H_1 \text{ if }\qquad |T_*| &> t_{1-\alpha/2,\,\nu_*} \\
               \sqrt{k}\cdot |T| &>t_{1-\alpha/2,\,k\nu} \\
               |T| &> \frac{1}{\sqrt{k}} t_{1-\alpha/2,\,k\nu}
\end{aligned}\end{equation}
which is a much less strict requirement on $T$ since
\begin{equation}
    t_{1-\alpha/2,\,\nu} > t_{1-\alpha/2,\,k\nu} > \frac{1}{\sqrt{k}} t_{1-\alpha/2,\,k\nu}
\end{equation}
and thus even previously insignificant gaps $\bar x_1 - \bar x_2$ can now be flagged as significant.

\subsection{Effect of large sample size on regression $t$-tests}\label{apx:largesample}
Most regressions by \citet{arnett2025why} are performed on at most a hundred measurements. In \Hone, however, a regression is done with one measurement \emph{per word}:
\begin{linequote}
    We fit a linear regression with number of tokens per word, word length in characters, and morphological types as predictors for MorphScore. We found that fertility and word length are both negatively correlated with MorphScore ($\chi^2(1)=61.457$, $p<0.001$; $\chi^2(1) = 364.03$, $p<0.001$; respectively); however, the effect sizes were extremely small with an adjusted $R^2 = 0.021$.
\end{linequote}
The response is modeled poorly ("small effect size") even though the hypothesis tests are highly significant. The reason for this is that the $t$-test\footnote{For high $n$, $t(\nu)^2 \to \chi^2(1)$, so test statistics can either be reported as being $T$-distributed or $\chi^2$-distributed.} statistic that checks whether a regression coefficient is significantly different from 0, is proportional to $\sqrt{n}$ \citep[p.\ 140]{seber_straight-line_2003}. The $p$-values are therefore explained by the sample size of $n = 23\,952$.

The reason why the response is not modeled properly is likely due to heavily skewed predictors: the data published by \citeauthor{arnett2025why} show that in almost all MorphScore languages, the token amount $M$ is distributed geometrically ($P(M)$ is geometric) while the word length $L$ is distributed binomially when keeping the token amount fixed ($P(L\mid M)$ is binomial). This means words with few tokens completely swamp words with many tokens, and to the regression, it looks like $M$ is basically a constant while $L$ is unrelated to it and binomially distributed. 

Given the negligible modeling performance and the fact that across MorphScore languages, the correlation between $\bar L\mid M$ and $M$ is around 77\%, the sign of the regression coefficients should not be relied on to infer the signs of the correlations between the predictors and the response.

\subsection{Causation does not imply correlation}\label{apx:causality}
In \Hone, a linear regression is performed that results in a poor model. The conclusion is that the tested predictors cannot explain the response:
\begin{linequote}
    We fit a linear regression with number of tokens per word, word length in characters, and morphological types as predictors for MorphScore. We found that fertility and word length are both negatively correlated with MorphScore ($\chi^2(1)=61.457$, $p<0.001$; $\chi^2(1) =$ $364.03$, $p<0.001$; respectively); however, the effect sizes were extremely small with an adjusted $R^2 = 0.021$.\\ \underline{Given these small effects}, \underline{lon}g\underline{er words or} \underline{hi}g\underline{her} \underline{fertilit}y \underline{cannot ex}p\underline{lain the }g\underline{reater }\underline{than $20\%$ hi}g\underline{her Mor}p\underline{hScores} \underline{for a}gg\underline{lutinative lan}g\underline{ua}g\underline{es}. -- \citeauthor{arnett2025why}
\end{linequote}
The underlying assumption here is that \textsl{causation implies correlation}: that is, if a predictor explains a response, then it should linearly correlate with it. By the \emph{modus tollens}, given that no correlation (linear relationship) is found -- indicated by $R^2$, not by the $p$-value of the coefficients -- it is assumed that the predictors do not explain the response. But causation does not imply correlation.

\paragraph{Example.} We sample $N = 20\,000$ values from a random uniform variable $X\sim U(-1+\Delta;\,1-\Delta)$ with $\Delta=0.04$ (so, 20 000 random numbers between -0.96 and +1.04). We then square each of them and call the result $Y$, so that $Y = X^2$. We now perform a linear regression $Y =\beta_0 + \beta_1 X +\varepsilon$. Despite the perfectly deterministic, perfectly explanatory, causal relationship between predictor and response, we get an adjusted $R^2 = 0.021$. (And here too, the coefficient $\beta_1 = 0.076$ is significantly different from 0 with $t(19\,998) = 20.760$, or equivalently $\chi^2(1)\approx 400$, $p < 0.001$.)

\newpage
\onecolumn
\section{Experimental Setup}\label{apx:setup}
\begin{table*}[ht]
	\centering
	\small
	\resizebox{\linewidth}{!}{
\begin{tabular}{lllp{0.2\linewidth}}
    \toprule
    Dataset & Link & $|$L$|$ & Language Coverage\newline (ISO-639-3) \\
    \midrule
    EuroParl \cite{koehn2005europarla,tiedemann2012parallel} & \href{https://huggingface.co/datasets/Helsinki-NLP/europarl}{huggingface.co/datasets/Helsinki-NLP/europarl} & 21 & bul, ces, dan, deu, ell, eng, est, fin, fra, hun, ita, lav, lit, nld, pol, por, ron, slk, slv, spa, swe\\
    Morpho Challenge \cite{kurimo2010morpho} & \href{https://morpho.aalto.fi/events/morphochallenge2010/datasets.shtml}{morpho.aalto.fi/events/morphochallenge2010/datasets} & 3 & eng,fin,tur\\
    MorphyNet \cite{batsuren2021morphynet} & \href{https://github.com/kbatsuren/MorphyNet}{github.com/kbatsuren/MorphyNet} & 15 & cat, ces, deu, eng, fin, fra, hbs, hun, ita, mon, pol, por, rus, spa, swe \\
    MorphScore \cite{arnett2025why} & \href{https://github.com/catherinearnett/morphscore}{github.com/catherinearnett/morphscore} & 22 & bul, ceb, ell, eng, eus, gle, guj, hun, hye, ind, isl, jpn, kat, kmr, kor, pes, slv, spa, tam, tur, urd, zul\\
    FineWeb 1 \& 2 \cite{penedo2024fineweb,penedo2025fineweb2} & \makecell[l]{\href{https://huggingface.co/datasets/HuggingFaceFW/fineweb}{huggingface.co/datasets/HuggingFaceFW/fineweb} \\ \href{https://huggingface.co/datasets/HuggingFaceFW/fineweb-2}{huggingface.co/datasets/HuggingFaceFW/fineweb-2}} & 63 & afr, amh, arb, bel, ben, bos, bul, cat, ces, cym, dan, deu, ell, eng, epo, eus, fin, fra, glg, guj, heb, hin, hrv, hun, hye, ind, isl, ita, jpn, kan, kat, kaz, kir, kor, lat, lit, mal, mar, mkd, mlt, nld, nob, pan, pol, por, ron, rus, sin, slk, slv, som, spa, srp, swe, tam, tat, tel, tgk, tha, tur, ukr, urd, vie\\
    Example sentence \autoref{tab:ppl} & \href{https://tatoeba.org/en/sentences/show/13142837}{tatoeba.org/en/sentences/show/13142837} & - & - \\
    \bottomrule
\end{tabular}
}
	\caption{Datasets used in our analysis. For FineWeb we report the languages we use, not all available languages.}
	\label{tab:datasets}
\end{table*}

\ifx\review\undefined
	\begin{table*}[h]
		\centering
		\small
		\resizebox{\linewidth}{!}{\begin{tabular}{lll}
    \toprule
    Library & Purpose & Link\\
    \midrule
    TkTkT & Tokenization toolkit. & \href{https://github.com/bauwenst/TkTkT}{github.com/bauwenst/TkTkT} \\ 
    MoDeST & Morphological datasets. & \href{https://github.com/bauwenst/MoDeST}{github.com/bauwenst/MoDeST}\\
    Qwanqwa & Language metadata. & \href{https://github.com/WPoelman/qwanqwa}{github.com/WPoelman/qwanqwa} \\
    Fugashi \cite{mccann2020fugashi} & Japanese word segmentation. & \href{https://github.com/polm/fugashi}{github.com/polm/fugashi} \\
    PyThaiNLP \cite{phatthiyaphaibun2023pythainlp} & Thai word segmentation. & \href{https://github.com/PyThaiNLP/pythainlp}{github.com/PyThaiNLP/pythainlp} \\
    -- & Scripts by \citet{arnett2025why}. & \href{https://osf.io/jukzd/?view_only=3d0d491d24074215a0ab81f72a693c16}{osf.io/jukzd} \\
    -- & Scripts used for this paper. & \href{https://github.com/LAGoM-NLP/ConfoundingFactors}{github.com/LAGoM-NLP/ConfoundingFactors} \\ 
    \bottomrule
\end{tabular}
}
		\caption{Software used in our analysis.}
		\label{tab:software}
	\end{table*}
\fi

\begin{table*}[h!]
    \centering
    \small
    \begin{tabular}{llr}
\toprule
Language & Source & $|$Words$|$ \\
\midrule
Turkish &  MorphoChallenge & 1000 \\
Mongolian &   MorphyNet & 16822 \\
Polish &   MorphyNet & 58711 \\
Swedish &   MorphyNet & 94488 \\
Catalan &  MorphyNet & 121749 \\
German &  MorphyNet & 186395 \\
Portuguese &  MorphyNet & 262116 \\
Czech &  MorphyNet & 326147 \\
French &  MorphyNet & 383064 \\
English &  MorphyNet & 414967 \\
Italian &  MorphyNet & 562273 \\
Russian &  MorphyNet & 804279 \\
Spanish &  MorphyNet & 882688 \\
Hungarian &  MorphyNet & 939067 \\
Finnish & MorphyNet & 1629510 \\
 \bottomrule
\end{tabular}

    \caption{Source (MorphoChallenge \citealt{kurimo2010morpho} or MorphyNet \citealt{batsuren2021morphynet}) and size of the morphological alignment datasets we use. Note that there can be multiple segmentation boundaries per word.}
    \label{tab:morph-dataset-stats}
\end{table*}

\newpage
\section{Full Results}\label{apx:results}
\begin{table*}[h!]
    \centering
    \resizebox{!}{0.45\textheight}{\begin{tabular}{ll|rr|rr|rr|rr|rr|rr|rr}
\multicolumn{2}{c|}{} & \multicolumn{4}{c|}{Token Bigrams} & \multicolumn{6}{c|}{Token Unigrams} & \multicolumn{4}{c}{Words} \\
\multicolumn{2}{c|}{} & \multicolumn{2}{c|}{AV} & \multicolumn{2}{c|}{$\eta$} & \multicolumn{2}{c|}{MATTR} & \multicolumn{2}{c|}{MTL} & \multicolumn{2}{c|}{RE} & \multicolumn{2}{c|}{$\mathcal S$} & \multicolumn{2}{c}{MWL} \\
\multicolumn{1}{l}{Language} & \multicolumn{1}{l|}{G$^*$} & \multicolumn{1}{c}{EP} & \multicolumn{1}{c|}{FW} & \multicolumn{1}{c}{EP} & \multicolumn{1}{c|}{FW ($\downarrow$)} & \multicolumn{1}{c}{ EP} & \multicolumn{1}{c|}{ FW} & \multicolumn{1}{c}{EP} & \multicolumn{1}{c|}{FW} & \multicolumn{1}{c}{EP} & \multicolumn{1}{c|}{FW} & \multicolumn{1}{c}{ EP} & \multicolumn{1}{c|}{ FW} & \multicolumn{1}{c}{EP} & \multicolumn{1}{c}{FW} \\
\hline
Thai & Iso &  & \gradproxy[3.370453995134856][8.953134742188528][20.33549408519737]{3.67} &  & \gradproxy[19.70983692317072][41.951147926523255][55.36589231511255]{19.71} &  & \badproxy[31.29273458862818][45.5984486616273][57.93295802312317]{52.16} &  & \badproxy[1.7731557725417246][4.308988134020461][5.780930620642329]{4.65} &  & \badproxy[24.40446412766168][36.08961265971455][54.21308748618633]{24.40} &  & \wordproxy[1.96339547959383][12.56724310960107][49.49930946927555]{3.69} &  & \wordproxy[1.9461971169397976][5.683411072659402][8.561332131993879]{5.05} \\
Vietnamese & Iso &  & \gradproxy[3.370453995134856][8.953134742188528][20.33549408519737]{3.37} &  & \gradproxy[19.70983692317072][41.951147926523255][55.36589231511255]{26.44} &  & \badproxy[31.29273458862818][45.5984486616273][57.93295802312317]{34.88} &  & \badproxy[1.7731557725417246][4.308988134020461][5.780930620642329]{3.78} &  & \badproxy[24.40446412766168][36.08961265971455][54.21308748618633]{40.31} &  & \wordproxy[1.96339547959383][12.56724310960107][49.49930946927555]{7.98} &  & \wordproxy[1.9461971169397976][5.683411072659402][8.561332131993879]{4.17} \\
\end{tabular}
}
    \caption{Full results with pretokenization. EP = EuroParl; FW = FineWeb. All metrics except for AV, MTL, and MWL are multiplied by 100 for visual clarity. $^*$The groupings are taken from \citeauthor{arnett2025why}.}
    \label{tab:full-results}
\end{table*}

\begin{table*}[ht]
    \centering
	\begin{subtable}{0.45\textwidth}
        \centering
        \resizebox{!}{0.46\textheight}{\begin{tabular}{ll|rr|rr}
\multicolumn{2}{c|}{} & \multicolumn{2}{c|}{AV} & \multicolumn{2}{c}{$\eta$} \\
\multicolumn{1}{l}{Language} & \multicolumn{1}{l|}{G$^*$} & \multicolumn{1}{c}{EP} & \multicolumn{1}{c|}{FW} & \multicolumn{1}{c}{EP} & \multicolumn{1}{c}{FW ($\downarrow$)} \\
\hline
Thai & Iso &  & \gradproxy[3.370453995134856][8.953134742188528][20.33549408519737]{3.67} &  & \gradproxy[19.70983692317072][41.951147926523255][55.36589231511255]{19.71} \\
Vietnamese & Iso &  & \gradproxy[3.370453995134856][8.953134742188528][20.33549408519737]{3.37} &  & \gradproxy[19.70983692317072][41.951147926523255][55.36589231511255]{26.44} \\
Indonesian & Agg &  & \gradproxy[3.370453995134856][8.953134742188528][20.33549408519737]{6.01} &  & \gradproxy[19.70983692317072][41.951147926523255][55.36589231511255]{29.55} \\
Galician & Fus &  & \gradproxy[3.370453995134856][8.953134742188528][20.33549408519737]{9.74} &  & \gradproxy[19.70983692317072][41.951147926523255][55.36589231511255]{30.80} \\
French & Fus & \gradproxy[2.1211252048993914][4.044892599519379][7.137534477729758]{2.39} & \gradproxy[3.370453995134856][8.953134742188528][20.33549408519737]{6.13} & \gradproxy[15.91942205034592][26.33009085190066][40.31395203003415]{19.11} & \gradproxy[19.70983692317072][41.951147926523255][55.36589231511255]{32.35} \\
Portuguese & Fus & \gradproxy[2.1211252048993914][4.044892599519379][7.137534477729758]{3.06} & \gradproxy[3.370453995134856][8.953134742188528][20.33549408519737]{6.79} & \gradproxy[15.91942205034592][26.33009085190066][40.31395203003415]{21.31} & \gradproxy[19.70983692317072][41.951147926523255][55.36589231511255]{33.02} \\
\end{tabular}
}
        \caption{With pretokenization.}
    \end{subtable}
    \hspace{2em}
	\begin{subtable}{0.45\textwidth}
        \centering
        \resizebox{!}{0.46\textheight}{\begin{tabular}{ll|rr|rr}
\multicolumn{2}{c|}{} & \multicolumn{2}{c|}{AV} & \multicolumn{2}{c}{$\eta$} \\
\multicolumn{1}{l}{Language} & \multicolumn{1}{l|}{G$^*$} & \multicolumn{1}{c}{EP} & \multicolumn{1}{c|}{FW} & \multicolumn{1}{c}{EP} & \multicolumn{1}{c}{FW ($\downarrow$)} \\
\hline
Galician & Fus &  & \gradproxy[18.497944507447283][30.45766123438304][54.445774561321215]{18.50} &  & \gradproxy[46.60691811530599][70.56394746805897][78.38652038859419]{46.61} \\
Vietnamese & Iso &  & \gradproxy[18.497944507447283][30.45766123438304][54.445774561321215]{24.34} &  & \gradproxy[46.60691811530599][70.56394746805897][78.38652038859419]{57.51} \\
Bosnian & Fus &  & \gradproxy[18.497944507447283][30.45766123438304][54.445774561321215]{26.16} &  & \gradproxy[46.60691811530599][70.56394746805897][78.38652038859419]{58.62} \\
Esperanto & Agg &  & \gradproxy[18.497944507447283][30.45766123438304][54.445774561321215]{26.74} &  & \gradproxy[46.60691811530599][70.56394746805897][78.38652038859419]{62.21} \\
Punjabi & Agg &  & \gradproxy[18.497944507447283][30.45766123438304][54.445774561321215]{28.13} &  & \gradproxy[46.60691811530599][70.56394746805897][78.38652038859419]{62.22} \\
Macedonian & Fus &  & \gradproxy[18.497944507447283][30.45766123438304][54.445774561321215]{26.25} &  & \gradproxy[46.60691811530599][70.56394746805897][78.38652038859419]{62.87} \\
\end{tabular}
}
        \caption{Without pretokenization.}
    \end{subtable}
    
    \caption{AV and $\eta$ results with and without pretokenization. EP = EuroParl; FW = FineWeb. The $\eta$ results are multiplied by 100 for visual clarity. $^*$The groupings are taken from \citeauthor{arnett2025why}.}
    \label{tab:av-pretok}
\end{table*}

\begin{figure*}[ht]
	\centering
    \begin{subfigure}{\textwidth}
    	\centering
    	\includegraphics[width=\textwidth]{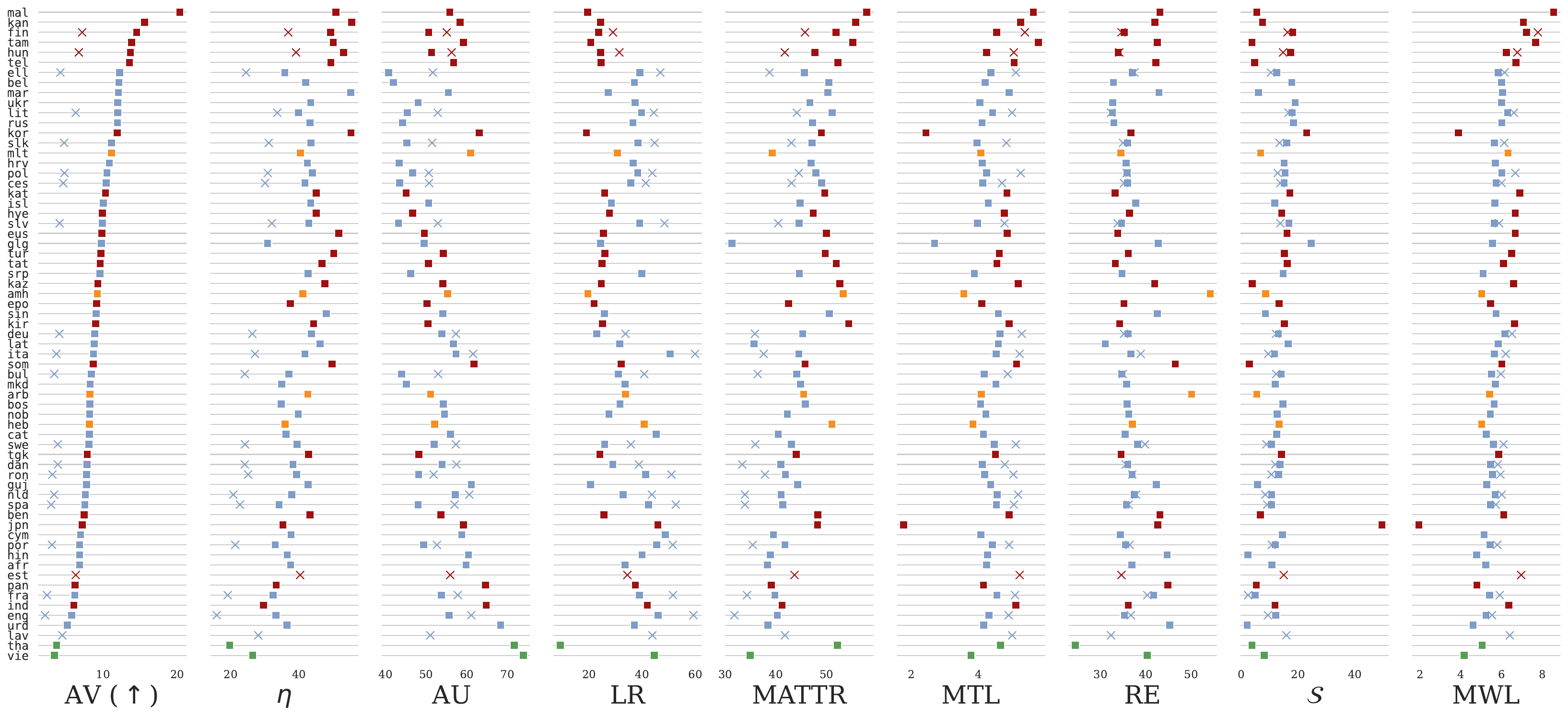}
    	\caption{Results sorted by AV. The bigram metrics are calculated \emph{with} pretokenization.}
    	\label{fig:with-pretok-av-sorted-results}
    \end{subfigure}
    \begin{subfigure}{\textwidth}
    	\centering
    	\includegraphics[width=\textwidth]{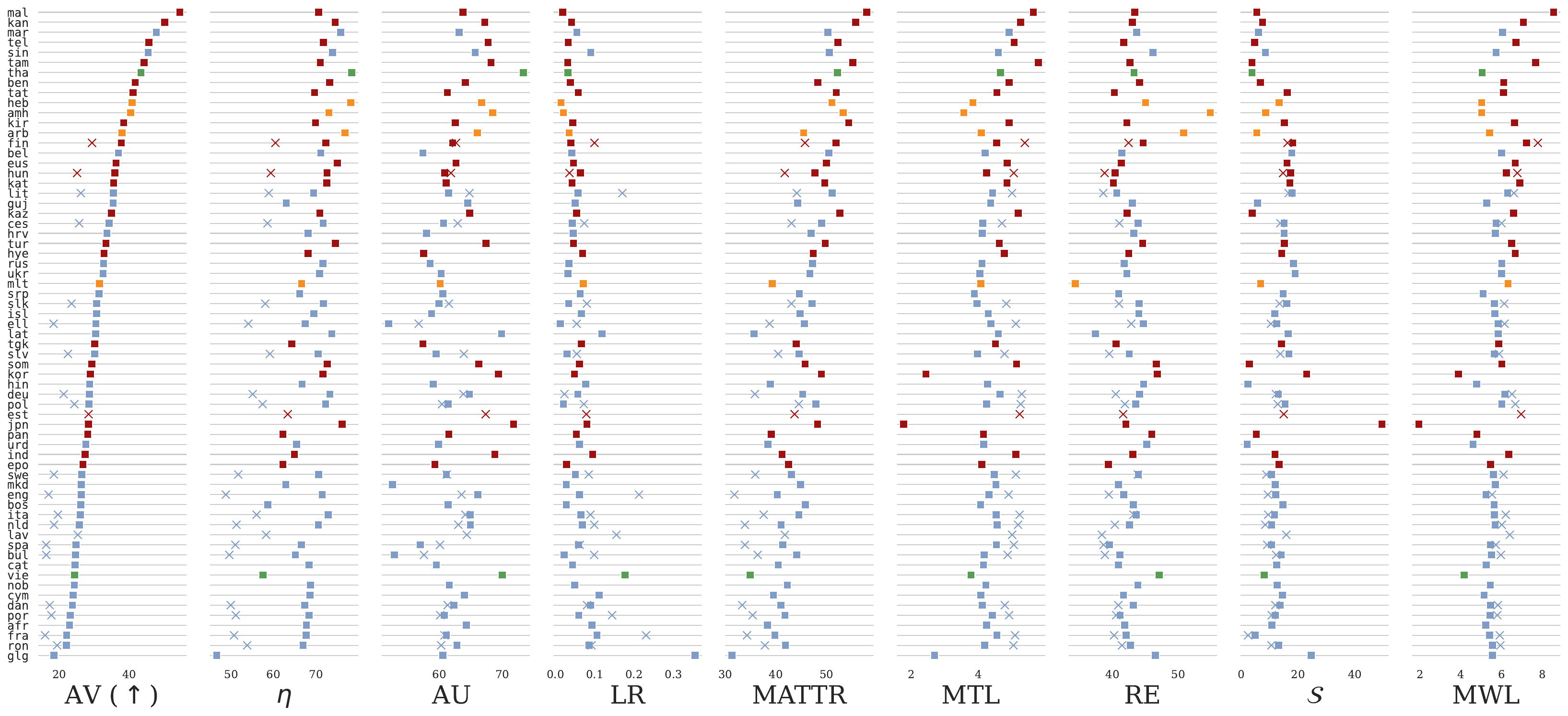}
    	\caption{Results sorted by AV. The bigram metrics are calculated \emph{without} pretokenization.}
    	\label{fig:no-pretok-av-sorted-results}
    \end{subfigure}
    \begin{subfigure}{\textwidth}
      \centering
	   \includegraphics[width=\textwidth]{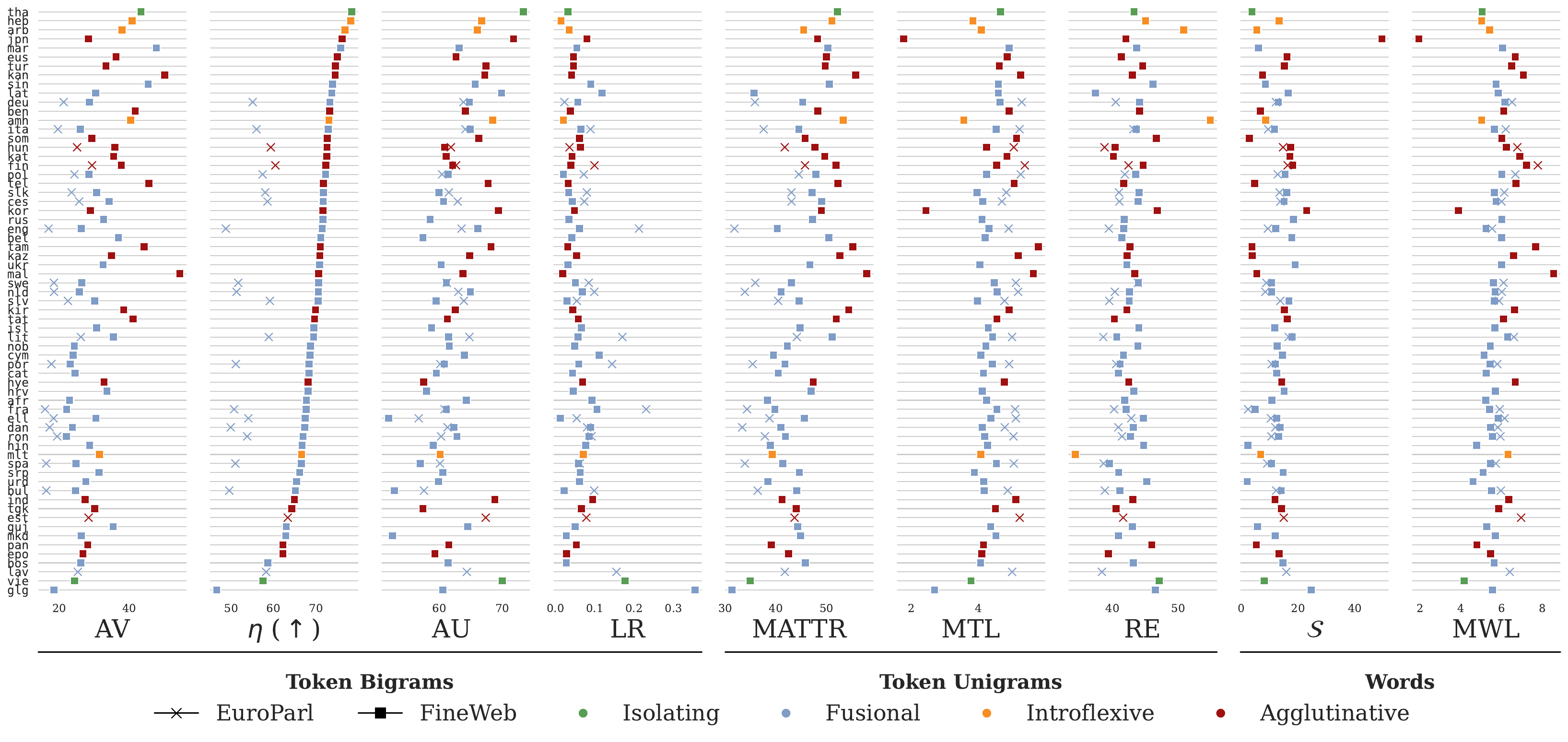}
	   \caption{Results sorted by $\eta$. The bigram metrics are calculated \emph{without} pretokenization.}
      \label{fig:no-pretok-eta-sorted-results}
    \end{subfigure}
    \caption{Only the sorting and pretokenization for the bigram metrics change, the other results are identical to \autoref{fig:results}.}
	\label{fig:additional-sorting-results}
\end{figure*}

\end{document}